\documentclass{article}
\usepackage{colm2024_conference}

\usepackage{booktabs}
\usepackage{graphicx}
\usepackage{amsmath}
\usepackage{enumitem}
\usepackage{wrapfig}
\usepackage{algorithm}
\usepackage{algpseudocode}
\usepackage{natbib}
\usepackage{makecell}
\usepackage{booktabs}
\usepackage{array}
\usepackage{amssymb}
\usepackage{amsfonts}
\usepackage{multirow}
\usepackage{verbatim}
\usepackage{caption}
\usepackage{longtable}
\usepackage{supertabular}
\usepackage{CJKutf8}
\usepackage[utf8]{inputenc}
\usepackage[T1]{fontenc}
\usepackage[french,vietnamese,mongolian,greek,english]{babel}
\usepackage{pifont}

\usepackage{enumitem}
\usepackage{tablefootnote}

\usepackage{xspace}
\usepackage{textcomp}
\usepackage{makecell}
\usepackage{lscape} 
\usepackage{siunitx}
\usepackage{listings}
\usepackage{xcolor}
\usepackage{svg}

\definecolor{dt}{gray}{0.7}
\usepackage{pifont}      
\usepackage{bbding}      
\usepackage{fontawesome}

\usepackage{scrextend}

\usepackage{tgpagella}
\usepackage{latexsym}
\usepackage[T1]{fontenc}
\usepackage[utf8]{inputenc}
\usepackage{microtype}
\definecolor{mydarkblue}{rgb}{0,0.08,0.45}
\definecolor{citecolor}{HTML}{0071BC}
\usepackage{url}           
\usepackage{nicefrac}       
\usepackage{changepage}
\usepackage{xargs}          
\usepackage{wrapfig,lipsum,booktabs}
\usepackage{longtable}
\usepackage{subcaption}
\usepackage{endnotes}

\usepackage{pgfplots}
\usetikzlibrary{pgfplots.groupplots}
\pgfplotsset{compat=1.3}
\usepackage{tikz}
\usetikzlibrary{patterns}

\usepackage{bm}

\usepackage[most]{tcolorbox}

\usepackage{hyperref}
\definecolor{darkblue}{rgb}{0, 0, 0.5}
\hypersetup{colorlinks=true, citecolor=darkblue, linkcolor=darkblue, urlcolor=darkblue, bookmarks=true, bookmarksopen=false, bookmarksnumbered=true}

\usepackage[capitalize,noabbrev]{cleveref}

\crefname{section}{\S}{\S\S}
\Crefname{section}{\S}{\S\S}
\crefname{subsection}{\S\S}{\S\S}
\Crefname{subsection}{\S\S}{\S\S}
\crefformat{section}{\S#2#1#3}
\crefformat{subsection}{\S\S#2#1#3}

\crefname{table}{Table}{Tables}
\crefname{figure}{Figure}{Figures}
\crefname{algorithm}{Algorithm}{}
\crefname{equation}{eq.}{}
\crefname{appendix}{Appendix}{}
\crefformat{section}{Section #2#1#3}
\usepackage{multicol}
\usepackage{tcolorbox}
\usepackage{titlesec}
\usepackage{float}
\titleformat*{\section}{\large\bfseries}
\usepackage{array} 
\newcolumntype{P}[1]{>{\centering\arraybackslash}p{#1}}
\usepackage{multirow}

\usepackage{adjustbox}
\definecolor{objblue}{RGB}{3,139,221}  
\definecolor{attrred}{RGB}{255,67,67}    
\definecolor{easygreen}{RGB}{0,156,75}  
\definecolor{middleyellow}{RGB}{242,89,34}  
\definecolor{hardred}{RGB}{216,56,58}

\usepackage{colortbl}
\usepackage{array}

\usepackage[most]{tcolorbox}
\definecolor{BoxBackground}{RGB}{240, 240, 240}
\definecolor{BoxFrame}{RGB}{0, 0, 0}
\definecolor{TitleBackground}{RGB}{0, 0, 0}
\definecolor{TitleText}{RGB}{255, 255, 255}

\tcbset{
  academicbox/.style={
    boxsep=5pt,
    left=2pt,
    right=2pt,
    bottom=0.5pt,
    boxrule=0.5pt,
    colback=BoxBackground,
    colframe=BoxFrame,
    colbacktitle=TitleBackground,
    coltitle=TitleText,
    enhanced,
    attach boxed title to top left={yshift=-0.1in,xshift=0.1in},
    boxed title style={boxrule=0pt,colframe=white},
    title={#1},
  }
}
\newtcolorbox{AcademicBox}[1][]{academicbox=#1}

\title{Qwen-Image-2.0 Technical Report}

\author{
\bf Qwen Team}

\begin{document}

\maketitle
\begin{abstract}
We present \textbf{Qwen-Image-2.0}, an omni-capable image generation foundation model that unifies high-fidelity image generation and precise image editing within a single integrated framework. While current image generation foundation models excel at high-quality aesthetic generation and text rendering, they still face significant challenges in practical creative workflows, including ultra-long text rendering, complex multilingual typography, high-resolution photorealism, robust instruction following, and efficient deployment. These limitations are particularly pronounced in text-rich and compositionally complex scenarios, where visual fidelity must be jointly maintained with semantic accuracy, typographic correctness, and layout coherence. More fundamentally, few existing systems can deliver all these capabilities for both image generation and image editing simultaneously within a single unified model without pipeline switching. To address these challenges, Qwen-Image-2.0 couples Qwen3-VL as the condition encoder with a Multimodal Diffusion Transformer for joint condition-target modeling, supported by comprehensive data curation and a customized multi-stage training pipeline. This design enables the model to leverage strong multimodal understanding while preserving the generative flexibility required for diverse creation and editing tasks. Specifically, Qwen-Image-2.0 enables ultra-long text rendering with instructions of up to 1K tokens, allowing direct generation of professional text-rich visual content such as slides, posters, infographics, and comics. It also substantially improves multilingual text rendering across diverse languages, with higher character fidelity and support for more complex and visually appealing typography. Beyond text-centric scenarios, the model advances high-resolution photorealistic image generation, producing richer local details, more realistic textures and materials, and more coherent lighting and shading. In addition, Qwen-Image-2.0 yields more stable quality across diverse artistic styles and follows complex prompts more faithfully, reducing concept omission, compositional failure, and hallucinated content. Extensive human evaluations show that Qwen-Image-2.0 delivers substantial improvements over previous Qwen-Image series models in both image generation and editing, demonstrating clear advances in overall visual quality, editing capability, and practical usability. We believe Qwen-Image-2.0 marks a meaningful step toward more general, reliable, and practical image generation foundation models, laying the groundwork for a unified generative backbone across contemporary visual creation, editing, and multimodal downstream applications.

\begin{figure}[ht]
    \centering
    \makebox[\linewidth]{
        \includegraphics[width=0.86\linewidth]{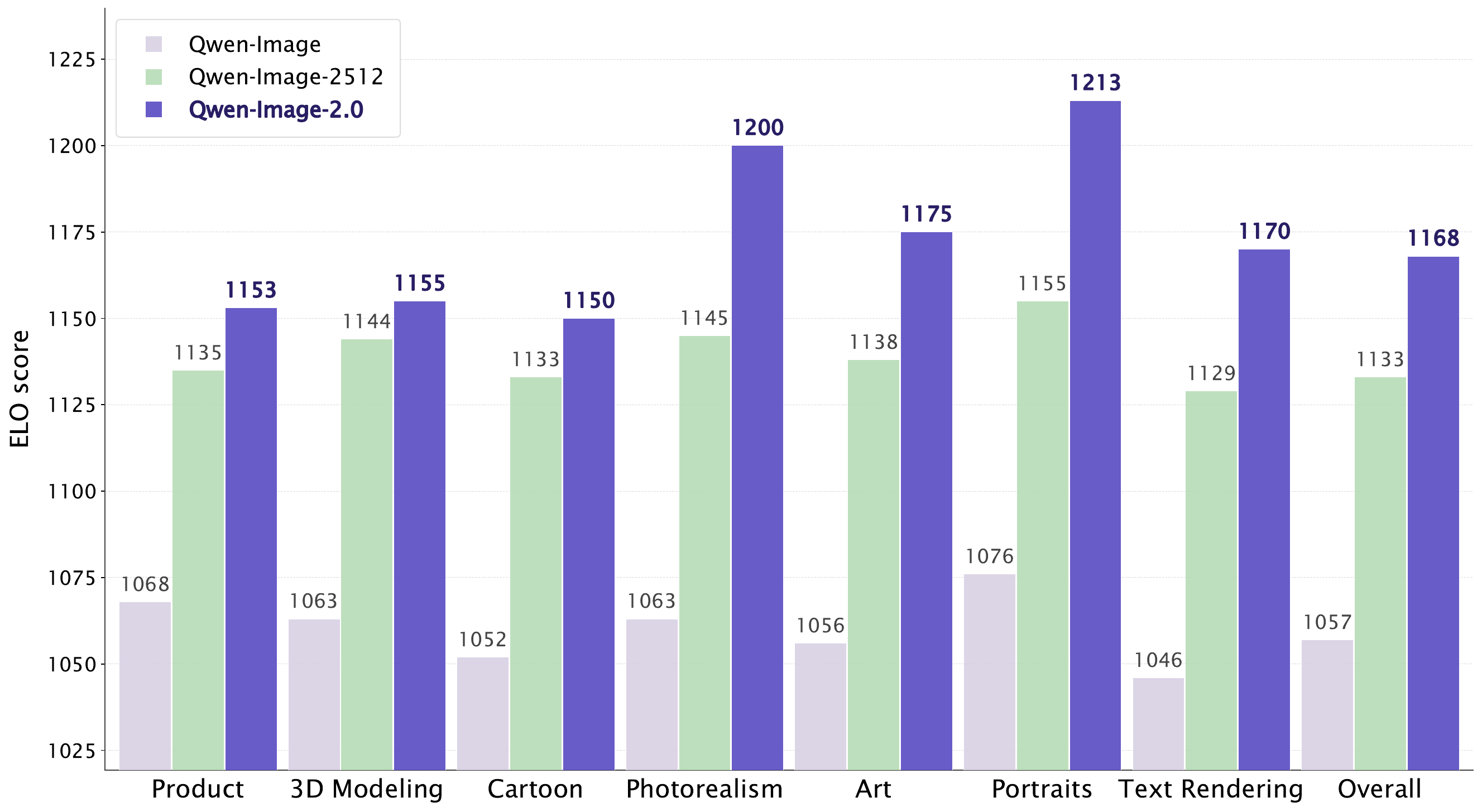}
    }
    \caption{Qwen-Image-2.0 shows significant improvements across core dimensions, including photorealism and portrait generation, in LMArena (accessed April 22, 2026).}
    \label{fig:lmarena_radar}
\end{figure}
\end{abstract}

\begin{figure}[H]
    \makebox[\linewidth]{
        \includegraphics[width=\linewidth]{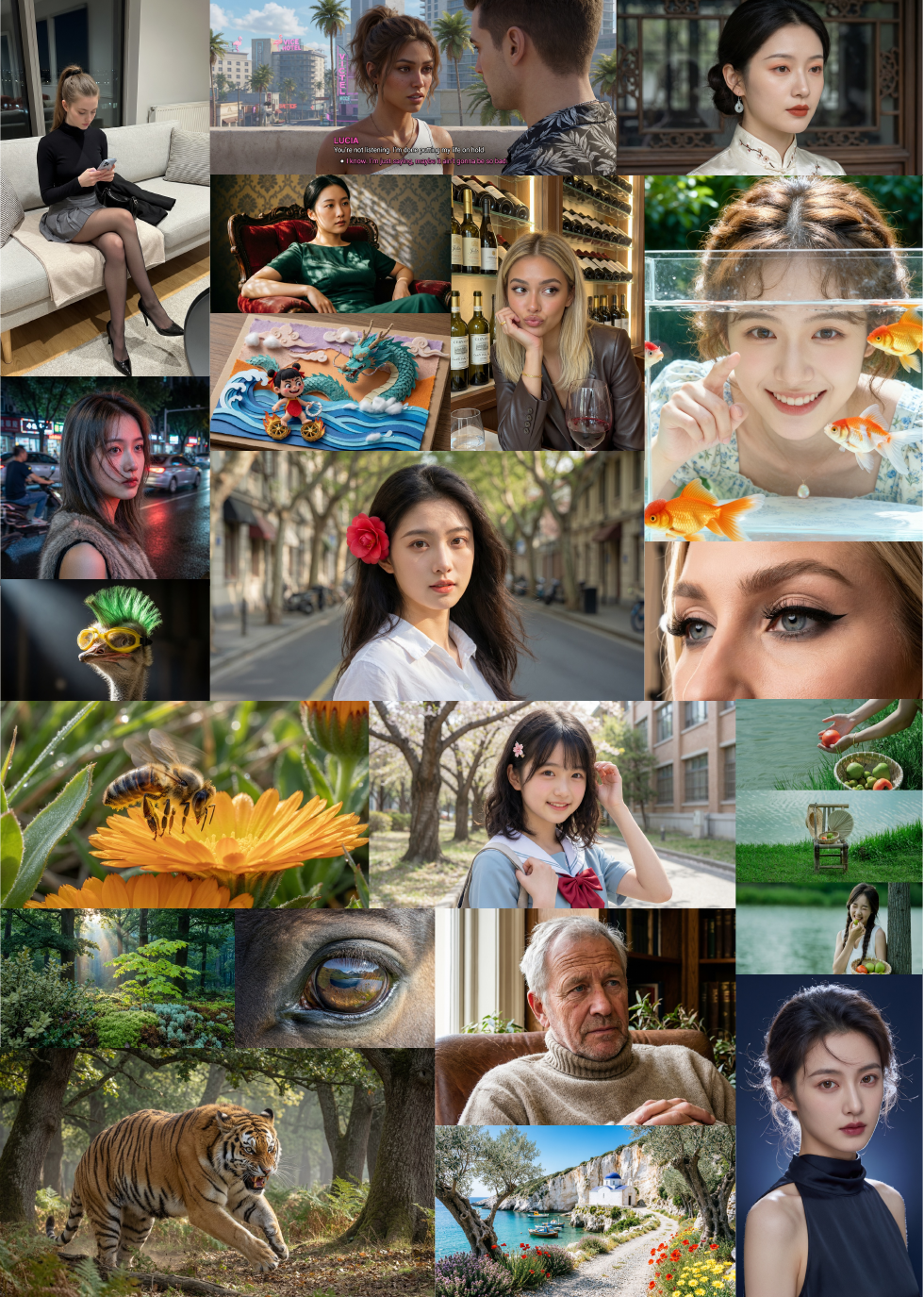}
    }
    \caption{Photo-realistic image generation showcase with Qwen-Image-2.0.}
\end{figure}

\begin{figure}[H]
    \makebox[\linewidth]{
        \includegraphics[width=\linewidth]{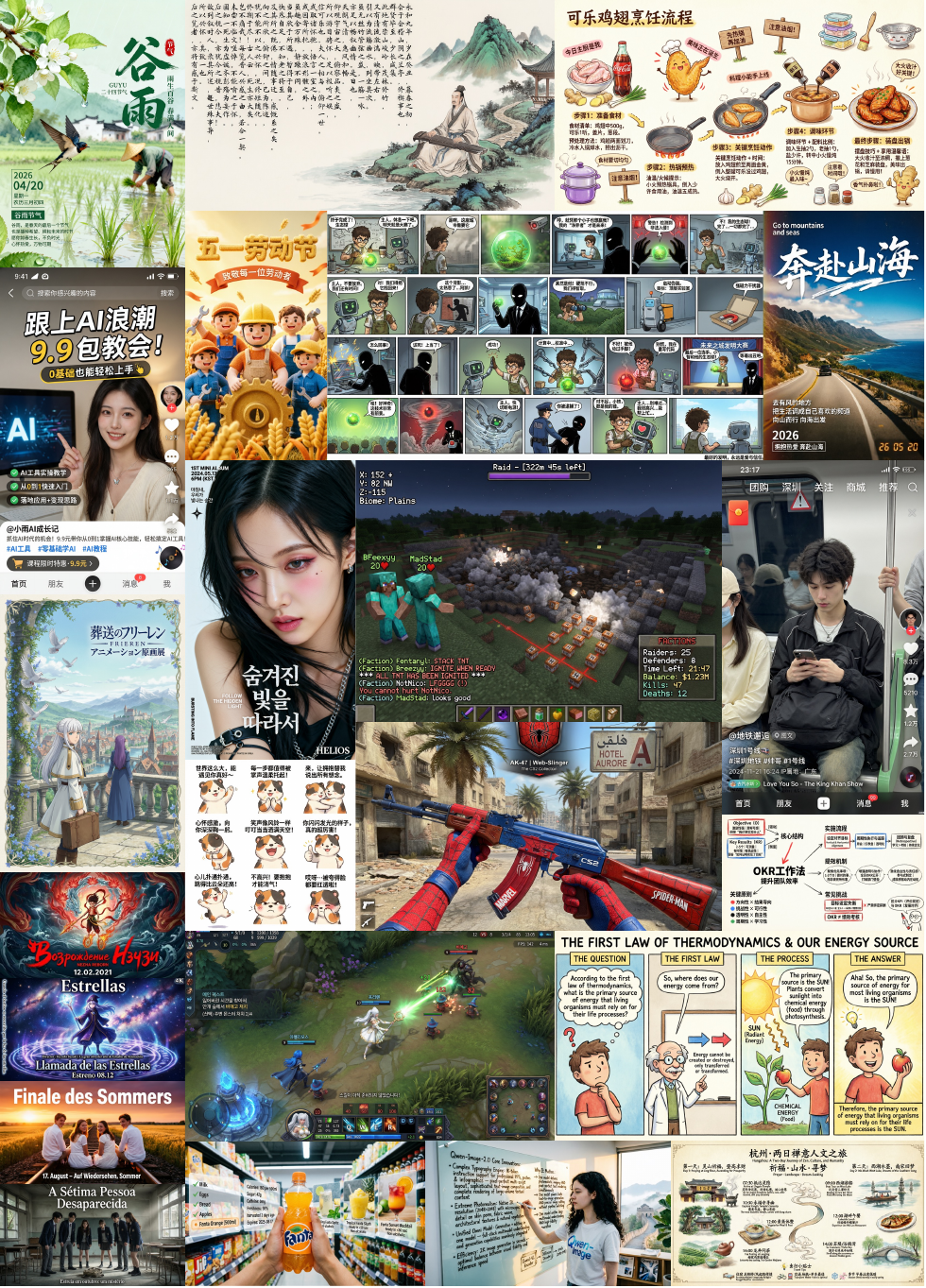}
    }
    \caption{Complex text rendering showcase with Qwen-Image-2.0.}
\end{figure}

\begin{figure}[H]
    \makebox[\linewidth]{
        \includegraphics[width=1\linewidth]{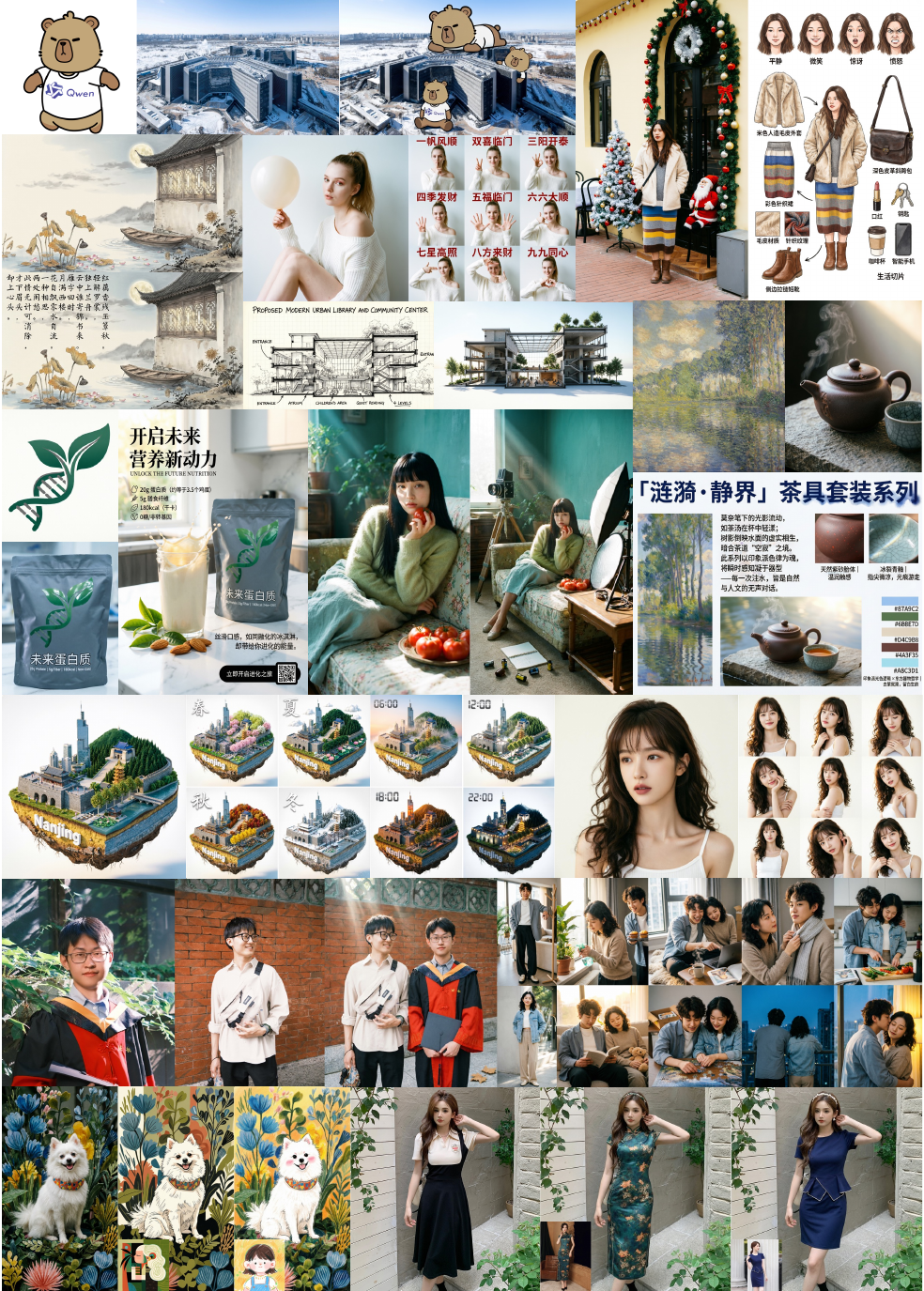}
    }
    \caption{Image editing showcase with Qwen-Image-2.0.}
\end{figure}

\newpage

\section{Introduction}

Image generation has progressed substantially, driven by the rapid advances in the research of multimodal foundation models~\citep{clip,Qwen2.5-VL,bai2025qwen3}. Diffusion and flow-based generative models~\citep{ho2020denoising,rombach2021highresolution,liu2022flow,lipman2022flow}, Transformer-based visual generation architectures~\citep{tian2024visual,han2025infinity,emu,imagegpt,yu2022scaling,chang2022maskgit}, and their advanced variants~\citep{peebles2023scalable,chen2024pixartalpha,esser2024scaling,ma2024sit} that combine the generative capacity of diffusion processes with the scalability of Transformer backbones have collectively established a powerful foundation for high-fidelity visual synthesis. The field has evolved from early latent diffusion models~\citep{rombach2021highresolution,podell2023sdxl} through diffusion Transformers~\citep{esser2024scaling,flux2024,flux-2-2025,labs2025kontext,li2024hunyuandit}, and more recent frameworks~\citep{wu2025qwen,team2025longcat,JoyAI-Image,hy-image-2.1,Cao2025HunyuanImage3T,cai2025z} have adopted vision-language foundation models as conditional encoders, whose stronger semantic grounding and multimodal world knowledge enable more precise instruction following and text-image alignment. Meanwhile, commercial systems~\citep{gao2025seedream,gong2025seedream,seedream2025seedream,seedream-5.0,gpt-image-1.5,nano-pro} have further pushed the frontier of generation quality and user experience. Together, these efforts have advanced the field to a point where high-fidelity image synthesis, visual text rendering, and instruction-based editing are becoming increasingly viable for real-world deployment.

Despite these progress, several bottlenecks persist when these models are deployed in real-world creative workflows. First, Ultra-long text rendering remains fragile: as the number of rendered characters grows, current models exhibit escalating glyph distortion, character omission, and layout collapse, limiting their utility for text-dense applications such as slides, infographics, and posters. Second, multilingual typography is underdeveloped; most systems are trained predominantly on English or Chinese glyphs and struggle to produce accurate characters, consistent spacing, or correct reading order for other scripts.
At higher resolutions, photorealistic generation also deteriorates---models often introduce repeated textures, incoherent lighting, and loss of fine-grained detail at 2K resolution and above, even when they can nominally produce large-canvas outputs.
For complex instruction following, prompts involving multiple entities, spatial constraints, or compositional logic frequently lead to concept omission or visual hallucination, revealing gaps in semantic understanding. Moreover, the computational cost of current architectures poses a significant efficiency bottleneck that constrains deployment in latency-sensitive and resource-limited settings.

Beyond these individual limitations, a more fundamental challenge lies in unifying these capabilities within a single model. Existing systems typically excel along one axis---producing either photorealistic imagery or accurate text rendering, supporting either text-to-image generation or image editing, but rarely deliver all capabilities simultaneously without resorting to separate pipelines or incurring notable quality trade-offs. Bridging deep multimodal understanding with high-fidelity generation for unifying text-to-image generation and image editing under a single, efficient architecture remains an open problem.

To address these challenges, we present \textbf{Qwen-Image-2.0}, an image generation foundation model that unifies text-to-image generation and image editing within a single framework.
Qwen-Image-2.0 is grounded in a comprehensive data infrastructure built around a fine-grained captioning framework tailored to different task types and image characteristics. A multi-stage, multi-resolution data pipeline progressively incorporates filtered corpora, editing pairs, synthetic data, and curated high-resolution samples, while an automated data flywheel leverages evaluation signals and user feedback to identify failure modes and drive iterative refinement.

Architecturally, the model couples a Qwen3-VL encoder~\citep{bai2025qwen3} with a Multimodal Diffusion Transformer (MMDiT, \citealt{esser2024scaling}) backbone. To enable native high-resolution generation, we introduce a high-compression Variational Autoencoder (VAE, \citealt{kingma2013auto}) with a 16$\times$ spatial downsampling ratio, incorporating residual autoencoding, enlarged latent channels, and a semantic alignment loss to balance compression efficiency, reconstruction fidelity, and latent diffusability. The MMDiT jointly models text and image tokens with MSRoPE~\citep{wu2025qwen} for cross-modal positional encoding, while using RMSNorm QK normalization, bias-free modulation, and SwiGLU activations to stabilize joint text-image training.

To bring these components together, we adopt a progressive multi-stage training recipe spanning large-scale pretraining, continual pretraining, supervised fine-tuning, and Reinforcement Learning from Human Feedback (RLHF). A resolution curriculum gradually scales from lower to higher resolutions, stabilizing optimization while improving detail fidelity and high-resolution coherence. For preference alignment, the RLHF stage uses task-specific reward models for aesthetics, text-image alignment, portrait quality, instruction following, and visual consistency, then optimizes the generation policy with a diffusion RL framework built on Group Relative Policy Optimization (GRPO, \citealt{liu2025flowgrpo,zheng2025diffusionnft}).

Together, these design choices yield a model that addresses the aforementioned bottlenecks in a unified architecture. The main contributions of Qwen-Image-2.0 are summarized as follows:
\begin{itemize}
    \item \textbf{Professional-grade text rendering with long-context support.} Qwen-Image-2.0 supports prompts of up to 1K tokens and can directly produce text-dense visual outputs such as slides, posters, and infographics, with substantially improved glyph fidelity over prior systems.
    \item \textbf{Broad multilingual rendering.} The model can handle a wide range of languages, with higher character accuracy and support for more beautiful and complex typography.
    \item \textbf{High-resolution photorealistic generation.} With native 2K-resolution support, Qwen-Image-2.0 produces finer texture detail, more coherent lighting, and more realistic materials across portraits, natural scenes, and architectural imagery.
    \item \textbf{Robust artistic expression across styles.} The model maintains robust quality under diverse aesthetic settings, effectively reducing quality fluctuation across artistic styles.
    \item \textbf{More precise instruction following.} Qwen-Image-2.0 demonstrates stronger semantic understanding for complex and composition-heavy prompts.
    \item \textbf{Unified generation and editing.} A single model supports both text-to-image generation and instruction-based image editing under a unified architecture and training paradigm.
    \item \textbf{Improved inference efficiency.} Through joint optimization of architecture and training strategy, Qwen-Image-2.0 achieves faster inference while preserving visual quality, making it well suited for interactive creative workflows.
\end{itemize}

\section{Data}
\label{sec:data}

\subsection{Data Collection}
\begin{figure}[ht]
\centering
\includegraphics[width=\linewidth, trim=30 0 30 0, clip]{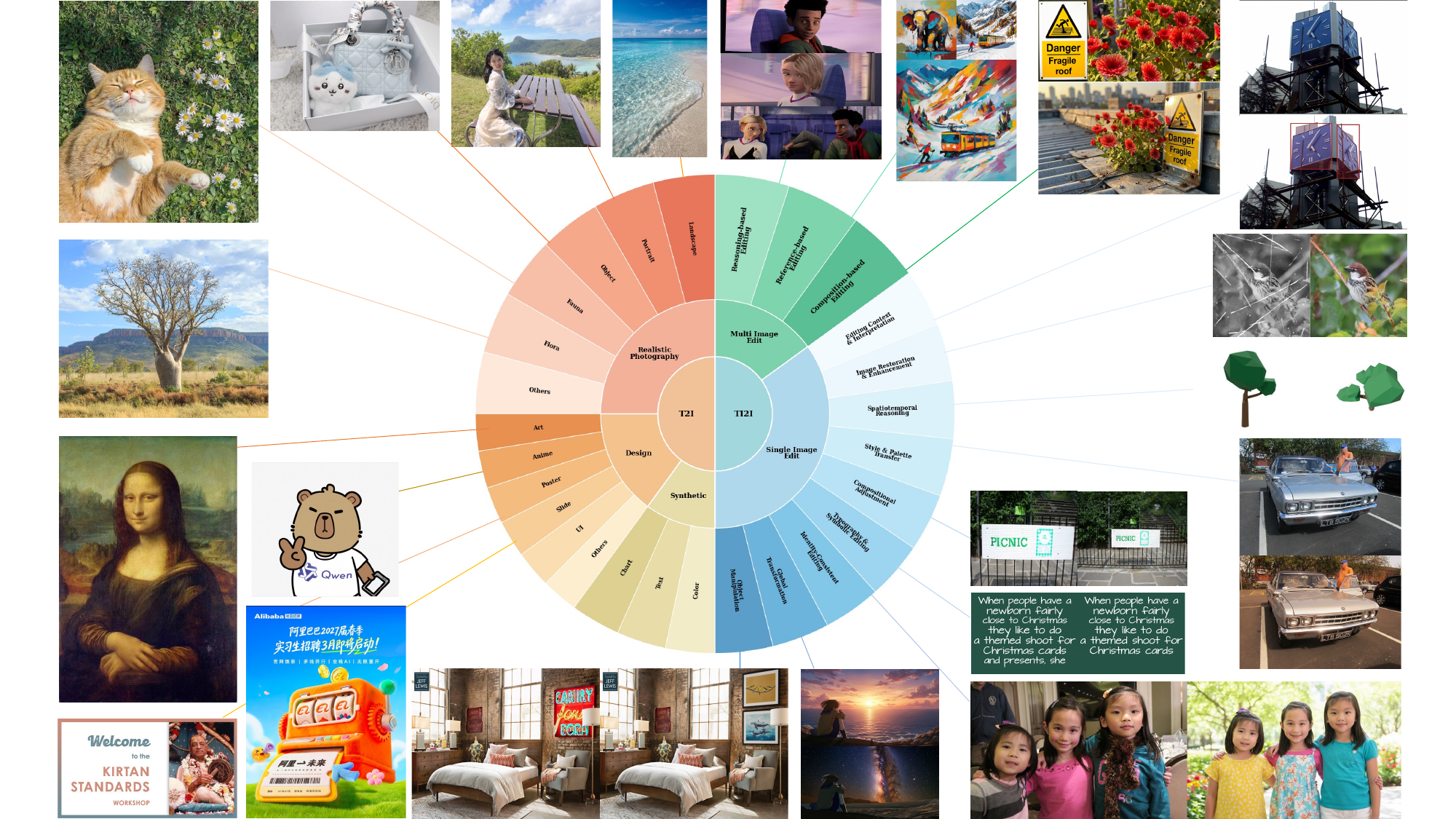}
   \caption{Qwen-Image-2.0 data distribution across image categories.}
\label{fig:data_distribution}
\end{figure}

We build a large-scale and diverse data pipeline to support unified training for both text-to-image generation and instruction-based image editing. Our data construction is guided by three principles: broad domain coverage, strong instruction quality, and reliable source-target consistency.

For Text-to-Image (T2I) generation, we collect image--text pairs spanning realistic photography, graphic design, artistic content, and synthetic imagery. The realistic subset covers common visual domains, including portraits, landscapes, objects, and other common visual domains, while preserving long-tail concepts and diverse scene compositions. Beyond natural images, we incorporate style-rich and layout-sensitive content, such as slides, posters, and rendered assets, to improve controllability over aesthetics, composition, and visual intent.

For image editing (TI2I), we curate and composite instruction-conditioned data in both single-image and multi-image settings. The single-image subset includes attribute modification, background replacement, style transfer, text editing, restoration, and structure-aware manipulation. The multi-image subset focuses on reference-based generation and editing, subject consistency, style transfer across images, and compositional merging. This coverage enables the model to learn a broad range of edit behaviors, from simple appearance changes to more complex transformations requiring semantic and spatial reasoning.

\subsection{Data Annotation}
To achieve comprehensive and detailed image descriptions across diverse and complex scenarios, we construct a fine-grained captioning framework tailored to different task types and image characteristics. Specifically, we design dedicated captioning schemes for General captions, Text captions, Knowledge captions, and Structured captions.

\paragraph{General captions}
General captions are designed for images of arbitrary resolution and complexity, aiming to provide comprehensive and detailed natural language descriptions of visual content.
This type covers not only the main objects, scene context, and spatial relationships in the image, but also textual content and its semantics whenever present. 
In addition, this type supports multilingual generation and varying caption lengths.

\paragraph{Text captions}
For images containing dense text or abstract symbols, we develop multiple prompting templates to specifically caption complex text-centric visual materials, such as presentation slides, comics, posters, educational materials, etc. 
Compared with general captions, this type place greater emphasis on accurately extracting dense textual content, layout structure, visual symbols, and their semantic relations.
As a result, this type is better suited for scenarios involving text-rich, structurally complex, and semantically organized images.

\paragraph{Knowledge captions}
Knowledge captions enrich the caption by injecting image-related background information, contextual cues, or auxiliary conditions in the form of conditions. 
This purpose is to enhance the model's ability to capture image semantics together with relevant world knowledge. 
Unlike captions that focus only on explicitly visible content, this type incorporates supplementary information associated with the image, helping the model build richer semantic connections and world knowledge.

\paragraph{Structured captions}
For images with complex relationships and numerous elements, such as relation graphs, flowcharts, and diagrams, natural language descriptions alone are often insufficient to fully and clearly represent the objects and their interactions. 
To address this issue, we adopt structured captions to explicitly model entities, attributes, and relations in the image. 
This type enables more accurate characterization of complex visual structures and facilitates the learning of hierarchical relations, topological dependencies, and semantic interactions among visual elements.

\subsection{Multi-Stage Training Data Strategy}
\begin{figure}[ht]
\centering
\includegraphics[width=1\linewidth, trim=0 160 0 30, clip]{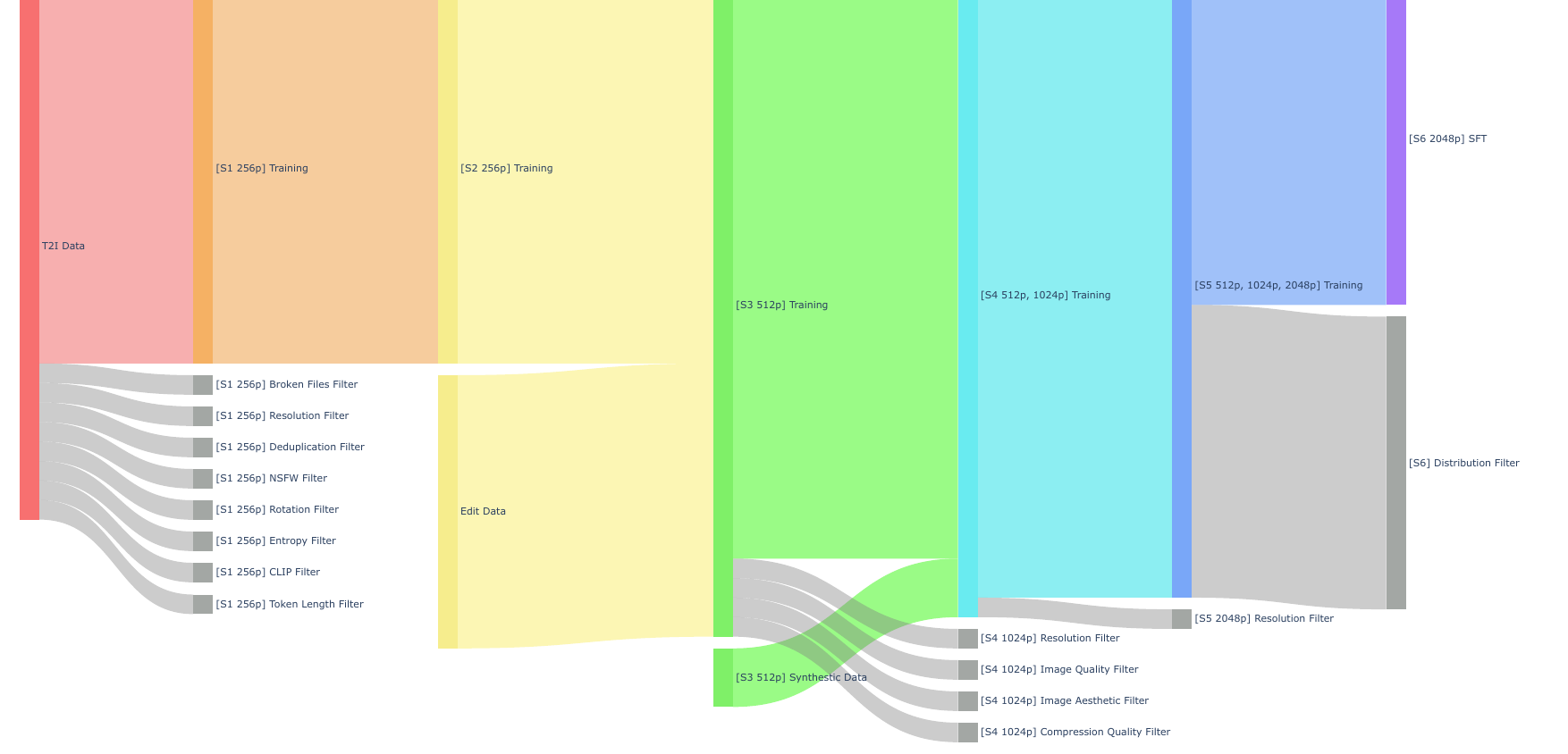}
   \caption{Overview of the Qwen-Image-2.0 data pipeline.}
\label{fig:pipeline}
\end{figure}

To ensure high-quality and well-curated training data throughout the iterative development of our
visual generation model. Based on Qwen-Image~\citep{wu2025qwen}, we designed a multi-stage filtering pipeline consisting of six sequential stages, as illustrated in Figure~\ref{fig:pipeline}. These filtering stages are applied progressively throughout the training process, with data distributions continuously refined over time.

\paragraph{Stage 1: 256P T2I pre-training} In the first stage, the raw T2I data undergoes a comprehensive set of eight sequential filters to establish a clean foundation for training. Since this stage targets training data at a 256$\times$ 256 resolution, we first apply a \textbf{Broken Files Filter} to remove corrupted or unreadable samples, followed by a \textbf{Resolution Filter} to discard images that cannot satisfy the required 256×256 resolution standard. A \textbf{Deduplication Filter} is then applied to eliminate redundant samples. Subsequently, a \textbf{NSFW Filter} removes inappropriate content, and a \textbf{Rotation Filter} corrects or discards images with improper orientations. An \textbf{Entropy Filter} is used to filter out images with abnormally low or high information content, and a \textbf{CLIP Filter} ensures strong image-text alignment by removing pairs with low similarity scores. Finally, a \textbf{Token Length Filter} removes samples whose text descriptions exceed the acceptable token length range.

\paragraph{Stage 2: 256P T2I \& TI2I pre-training}
Building upon the filtered 256p T2I data from Stage 1, Stage 2 introduces Edit Data to support text-guided image editing tasks. The filtered T2I data and TI2I data are combined and used directly for Stage 2 training. At this stage, all training is conducted at 256p resolution, enabling the model to learn both text-to-image generation and text-guided image editing under a unified low-resolution pre-training setting.

\paragraph{Stage 3: 512P T2I \& TI2I pre-training}
Stage 3 scales the training resolution from 256p to 512p. In addition to the data carried over from Stage 2, Synthetic Data is introduced to enrich the training distribution and improve data diversity at the higher 512p resolution. The combined dataset, consisting of filtered T2I data, Edit Data, and Synthetic Data, is then used for Stage 3 training, allowing the model to further improve its generation and editing capabilities at 512p.

\paragraph{Stage 4: 512P/1024P T2I \& TI2I pre-training}
Stage 4 further extends pre-training to a mixed-resolution setting covering both 512p and 1024p data. To support training at 1024p resolution, additional filtering steps are applied to ensure that the selected samples are suitable for high-resolution learning. Specifically, a \textbf{Resolution Filter} is used to retain images with sufficient spatial resolution, an \textbf{Image Quality Filter} removes low-fidelity images, an \textbf{Image Aesthetic Filter} selects visually appealing samples, and a \textbf{Compression Quality Filter} discards heavily compressed or artifact-laden images. The resulting high-quality 512p/1024p dataset is used for Stage 4 training.

\paragraph{Stage 5: Multi-Resolution T2I \& TI2I pre-training}
Stage 5 expands the training regime to a broader multi-resolution setting, covering 512p, 1024p, and 2048p resolutions. To support the newly introduced 2048p training data, a dedicated \textbf{Resolution Filter} is applied to select images that satisfy the stricter 2048p resolution requirement. This stage enables the model to learn from data across multiple scales and further strengthens its ability to generate and edit high-resolution images.

\paragraph{Stage 6: Supervised fine-tuning}
The final stage performs supervised fine-tuning (SFT) to better align the model with high-quality human preferences. Unlike the preceding pre-training stages, which progressively expand the resolution range from 256p to 2048p, Stage 6 focuses on refining the data distribution and sample quality across the target high-resolution settings. A \textbf{Distribution Filter} is applied to remove low-quality or imbalanced samples by reusing the filtering operators from previous stages with stricter thresholds. The refined data is then used for SFT, producing the final fine-tuned model optimized for high-resolution, high-fidelity visual generation and editing.

\subsection{Closed-loop Data Flywheel System}
\label{sec:data:data_feedback_system}

\begin{figure*}[t]
\centering
\includegraphics[width=\linewidth, trim=0 0.8cm 0 0, clip]{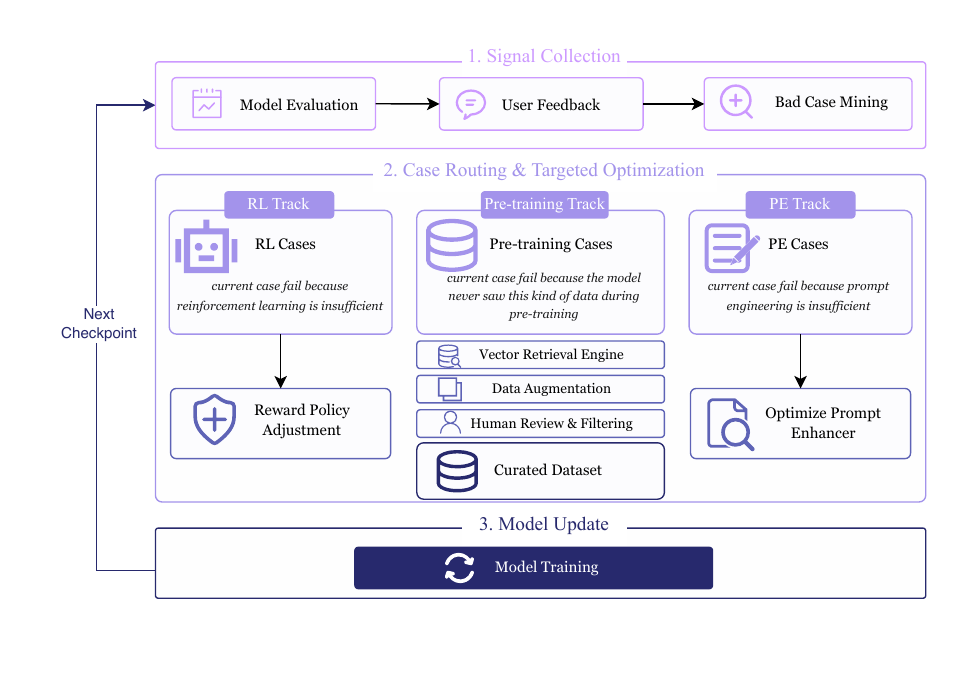}

\caption{
    An error-attribution-driven closed-loop data flywheel for multi-track targeted optimization.
}
\label{fig:data_feedback_system}
\end{figure*}

To continuously optimize the image generation and editing models and achieve iterative capability enhancement, we design and introduce a highly automated \textbf{Data Flywheel System}. As illustrated in Figure~\ref{fig:data_feedback_system}, this system comprises a closed loop consisting of three core stages:

\begin{itemize}
    \item \textbf{Stage 1: Multi-source signal collection.} The flywheel begins with a comprehensive assessment of the model's current capabilities and failure modes. The system automatically collects feedback signals through standardized model evaluation, targeted bad-case mining, and user feedback from diverse sources, including both real-world online interactions and internally self-evaluated cases generated during the training process. Together, these signals establish a robust data foundation for subsequent model optimization.
    
    \item \textbf{Stage 2: Case routing \& targeted optimization.} The collected failure cases are not processed in a uniform manner. Instead, they are automatically routed to three distinct optimization tracks according to an error attribution mechanism:
    \begin{itemize}
        \item \textbf{RL track.} For alignment or policy-related issues caused by insufficient reinforcement learning, the system assigns the corresponding cases to the RL track and addresses them through automated reward policy adjustment.
        \item \textbf{Pre-training track.} If a failure is attributed to missing knowledge, \textit{i.e.}, the model has not been sufficiently exposed to similar data during pre-training, the case is routed to the pre-training-oriented data compensation track. In this track, the system automatically invokes a vector retrieval engine with two objectives: first, to diagnose whether the failure is caused by the scarcity of specific data categories; and second, to retrieve and generalize diverse text prompts for image generation, as well as comprehensive instruction-image pairs for image editing, including editing prompts and their corresponding base images. Through automated data augmentation and the only manual intervention in the pipeline, namely necessary human review \& filtering, a curated dataset is constructed to bridge the identified knowledge gap.
        \item \textbf{Prompt engineering track.} When the model already possesses the required capability but fails due to inaccurate instruction understanding or suboptimal prompt formulation, the case is assigned to the prompt engineering track, where the system automatically refines the input through an optimized prompt enhancer.
    \end{itemize}
    
    \item \textbf{Stage 3: Model update \& closed loop.} After aggregating the strategies, new datasets, and parameter updates from the above tracks, the system automatically initiates the next training round. The resulting checkpoint is then fed back to Stage 1 for evaluation and deployment. This iterative process of ``failure discovery, targeted remediation, and model update'' forms a self-reinforcing optimization loop.
\end{itemize}

In summary, this data flywheel system provides a highly automated closed-loop framework for continuous model evolution. By limiting manual intervention to critical data filtering, it substantially reduces engineering overhead while preserving data reliability. Moreover, its error attribution mechanism enables targeted and resource-efficient optimization, while the vector retrieval engine continuously enriches the diversity of training data, thereby improving the model's generalization ability and robustness in complex generation and editing scenarios.

\section{Architecture}
\label{sec:model_arch}  

\begin{figure}[ht]
\centering
\includegraphics[width=0.86\linewidth]{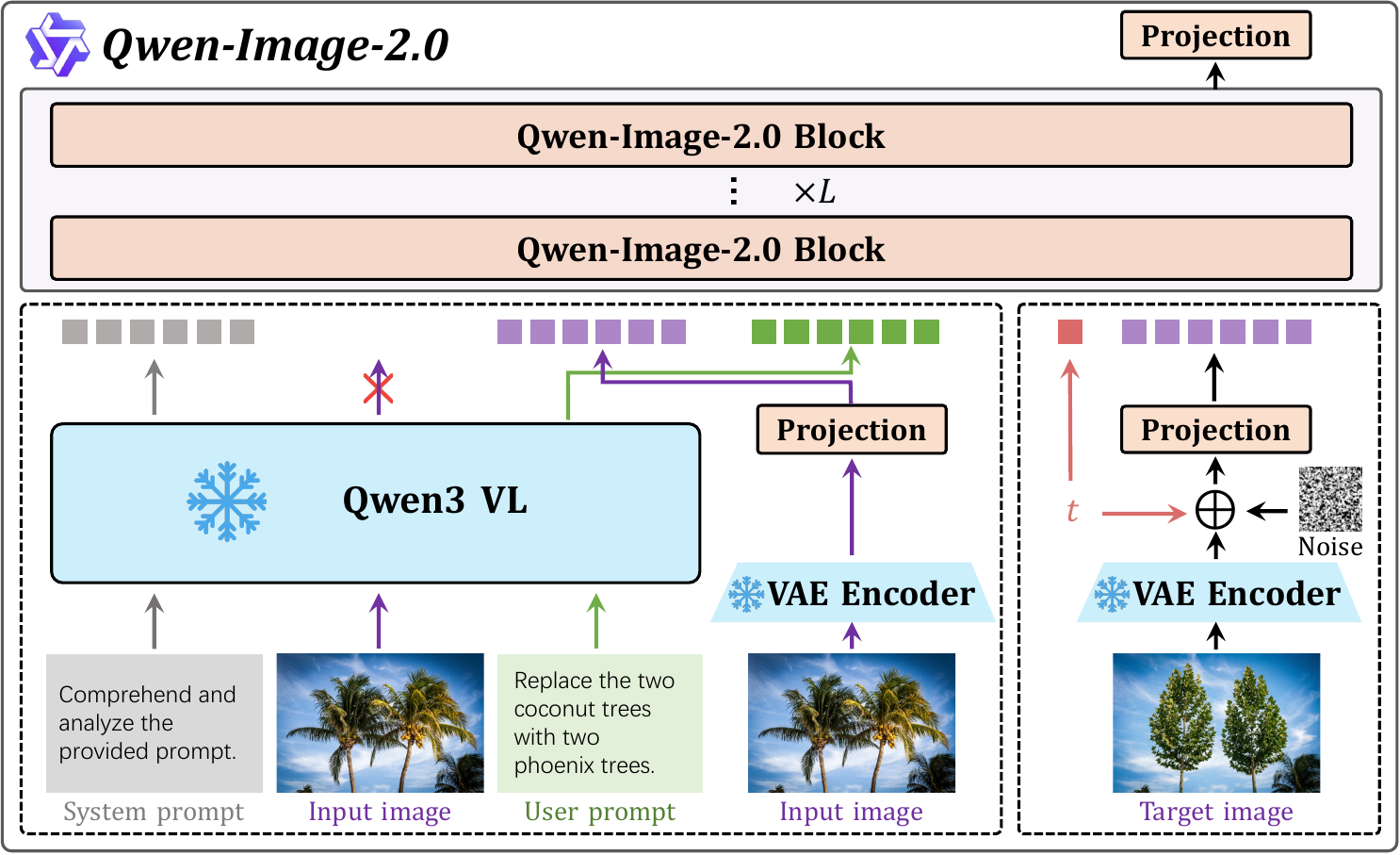}
   \caption{Overview of the Qwen-Image-2.0 architecture. The model adopts a MMDiT architecture, with input representations provided by a frozen Qwen3-VL and a VAE encoder. It uses RMSNorm~\citep{zhang2019root} for QK-Norm, while all other normalization layers use LayerNorm. The unified stream, comprising both text and image modalities, employs the joint positional calculation with MSRoPE encoding introduced in Qwen-Image~\citep{wu2025qwen}. SwiGLU is adopted as the non-linear activation function in the MLP layers to improve expressivity and enhance training stability.}
\label{fig:arc}
\end{figure}

As shown in Figure~\ref{fig:arc}, the Qwen-Image-2.0 architecture comprises three core, tightly coupled functional components that work in concert to enable high-fidelity, controllable, and efficient T2I generation. The first is a Multimodal Large Language Model (\textbf{MLLM}), instantiated as Qwen3-VL~\citep{bai2025qwen3} in our implementation, which serves as the condition encoder and extracts semantic features from user inputs. The second is a \textbf{VAE}, which encodes images into latent representations and decodes generated latents back into the image space. The third is a \textbf{MMDiT}, which performs the core denoising process in the latent space conditioned on the multimodal representations.

\subsection{Variational AutoEncoder}
\label{subsec:vae}
High-compression VAEs are crucial for native high-resolution image synthesis, as they substantially reduce diffusion training costs by projecting images into compact latent representations. Whereas existing open-source VAEs~\citep{wan2025wan, kong2024hunyuanvideo, wu2025qwen} typically adopt an 8$\times$ compression ratio, we employ a 16$\times$ ratio to further accelerate DiT training.

However, high-compression VAEs inevitably confront a three-way trade-off among compression ratio, reconstruction fidelity, and diffusability (\textit{i.e.}, the ease with which the latent space can be modeled by diffusion). On the one hand, aggressive compression introduces severe information bottlenecks, thereby compromising reconstruction quality. On the other hand, preserving information by increasing the number of latent channels yields high-dimensional latent manifolds that are difficult to diffuse, resulting in slower convergence and degraded generation quality.

To mitigate the reconstruction bottleneck, we adopt a residual autoencoder architecture~\citep{Chen2024DeepCA}, which incorporates non-parametric shortcut connections to better preserve fine-grained spatial details. In addition, we increase the latent dimensionality to 64 channels. This $f16c64$ configuration preserves the same total channel bottleneck as the standard $f8c16$ baseline, enabling high-fidelity reconstruction under a higher compression ratio. To further improve reconstruction quality in text-dense scenarios, we train the model on a large-scale internal corpus of text-rich images. The corpus includes real-world documents (\textit{e.g.}, PDFs, presentation slides, and posters) as well as synthetic paragraphs, covering both alphabetic scripts such as English and logographic scripts such as Chinese.

To enhance latent-space diffusability, we follow VA-VAE~\citep{Yao2025ReconstructionVG} and introduce a semantic alignment loss in addition to conventional reconstruction objectives. Specifically, we align the learned latent space with semantic representations over a broad image collection spanning diverse domains, aspect ratios, and resolutions. The VAE is optimized with reconstruction, perceptual, and semantic alignment losses. During optimization, we make two key observations. First, dynamic semantic alignment is highly effective: imposing strong semantic alignment constraints in early training is essential for establishing a diffusable latent space, while gradually relaxing this constraint later enables a better balance between reconstruction fidelity and diffusability. Second, adversarial loss is largely redundant in large-scale VAE training, consistent with recent findings~\citep{wu2025qwen}. We therefore remove the adversarial objective to improve training stability.

\paragraph{VAE reconstruction performance} \label{sec:vae_exp}
We quantitatively compare Qwen-Image-2.0-VAE with state-of-the-art image tokenizers using Peak Signal-to-Noise Ratio (PSNR) and Structural Similarity Index Measure (SSIM) as reconstruction metrics. Following prior work, we evaluate general-domain reconstruction on the ImageNet-1k~\citep{deng2009imagenet} validation set at 256$\times$256 resolution. To assess fidelity on small and dense text, we further report results on an in-house text-rich corpus~\citep{wu2025qwen} comprising diverse text sources and languages. As shown in Table~\ref{tab:table_vae}, Qwen-Image 2.0-VAE achieves state-of-the-art performance across all metrics under a 16$\times$ compression ratio.

\begin{table}[t]
\centering
\caption{Quantitative evaluation results of VAEs under different settings.}
\resizebox{\linewidth}{!}{
\begin{tabular}{lc|cc|cc|cc}
\toprule
\multirow{2}{*}{\textbf{Model}} & \multirow{2}{*}{\textbf{Setting}} & 
\multicolumn{2}{c|}{\textbf{\# Params (M)}} & 
\multicolumn{2}{c|}{\textbf{Imagenet\_256x256}} & 
\multicolumn{2}{c}{\textbf{Text\_256x256}} \\
\cmidrule(lr){3-4} \cmidrule(lr){5-6} \cmidrule(lr){7-8}
& & {Enc} & {Dec} & {PSNR} & {SSIM} & {PSNR} & {SSIM} \\
\midrule
SD-3.5 \citep{esser2024scaling} & f8c16 & 34 & 50 & 31.22 & 0.8839 & 29.93 & 0.9658 \\
Cosmos-CI8x8 \citep{agarwal2025cosmos} & f8c16 & 31 & 46 & 32.23 & 0.9010 & 30.62 & 0.9664 \\
Wan2.1 \citep{wan2025wan} & f8c16 & 54 & 73 & 31.29 & 0.8870 & 26.77 & 0.9386 \\
HunyuanVideo \citep{kong2024hunyuanvideo} & f8c16 & 100 & 146 & 33.21 & 0.9143 & 32.83 & 0.9773 \\
FLUX.1-dev \citep{flux2024} & f8c16 & 34 & 50 & 32.84 & 0.9155 & 32.65 & 0.9792 \\
\textbf{Qwen-Image} \citep{wu2025qwen} & f8c16 & 54 & 73 & \textbf{33.42} & \textbf{0.9159} & \textbf{36.63} & \textbf{0.9839} \\
\midrule
HunyuanImage-3.0 \citep{Cao2025HunyuanImage3T} & f16c32 & 389 & 871 & 31.08 & 0.8655 & 29.23 & 0.9521 \\
Wan2.2 \citep{wan2025wan} & f16c48 & 150 & 555 & 31.30 & 0.8784 & 28.19 & 0.9508 \\
Stepvideo-T2V \citep{Ma2025StepVideoT2VTR} & f16c64 & 110 & 389 & 31.54 & 0.8973 & 29.62 & 0.9641 \\
\textbf{Qwen-Image-2.0} & f16c64 & 79 & 259 & \textbf{33.42} & \textbf{0.9225} & \textbf{32.81} & \textbf{0.9795} \\
\bottomrule
\end{tabular}
}
\label{tab:table_vae}
\end{table}

\subsection{Multi-modal Diffusion Transformer}
\label{subsec:mmdit}
Figure~\ref{fig:arc} illustrates the overall architecture of Qwen-Image-2.0, a unified framework for T2I and TI2I generation that naturally supports interleaved multi-image inputs. To jointly and efficiently model textual and visual modalities, it adopts a MMDiT~\citep{esser2024scaling} architecture, where text and image tokens are processed within a shared transformer backbone.

Specifically, given visual inputs $\bm{x}$ and textual inputs $\bm{y}$, Qwen3-VL~\citep{bai2025qwen3} first encodes them into modality-aware representations $\bm{h}_{\bm{x}}$ and $\bm{h}_{\bm{y}}$, respectively. The visual representation $\bm{h}_{\bm{x}}$ is then replaced by the latent representation extracted by the variational autoencoder, denoted as $\mathcal{E}_{\bm{x}}$. The resulting multimodal sequence is constructed by concatenation:
\begin{equation}
    \bm{h} = \mathrm{Concat}\left(\mathcal{E}_{\bm{x}}, h_{\bm{y}}\right),
\end{equation}
which is subsequently fed into the Qwen-Image-2.0 block. To encode positional information across both textual and visual tokens in a unified manner, we employ MSRoPE~\citep{wu2025qwen} within the attention module. For the modulation module, we remove the bias term and adopt a purely multiplicative modulation formulation:
\begin{equation}
    \bm{h}^{\prime} = \alpha \bm{h},
\end{equation}
instead of the conventional affine form $\bm{h}^{\prime} = \alpha \bm{h} + \beta$, where $\alpha$ and $\beta$ denote scalar modulation parameters.

In practice, we observe that joint text-image training may induce excessively large activation magnitudes, leading to premature neuron saturation in the model~\citep{sunmassive}. To alleviate this issue, we introduce a SwiGLU module into the Multilayer Perceptron (MLP) layers. Given a latent representation $\bm{x}$, the SwiGLU transformation is formulated as
\begin{equation}
    \bm{h} = \Phi_{1}\left(\bm{x}\right) \otimes \sigma\left(\Phi_{2}\left(\bm{x}\right)\right),
\end{equation}
where $\Phi_{1}(\cdot)$ and $\Phi_{2}(\cdot)$ denote linear projection functions, $\sigma(\cdot)$ is the SiLU activation function, and $\otimes$ represents element-wise multiplication.

\subsection{Prompt Enhancer}
For complex image generation tasks, such as infographics, posters, typographic layouts, multi-panel storyboards, and data visualizations, generation quality depends on both the model's visual synthesis capacity and the prompt's specification of layout, object relations, visual hierarchy, and compositional intent. However, real-world user prompts vary substantially in granularity and explicitness, creating a key bottleneck for high-complexity visual creation. To this end, we introduce the Prompt Enhancer (PE), a rewriting module that converts user queries of varying specificity into structured, detail-rich prompts, enabling the downstream generator to better capture the intended visual design across diverse tasks.

\paragraph{Data Construction}
We construct prompt-enhancement data via a reverse-engineering pipeline that atomically degrades fine-grained annotations into diverse, colloquial user prompts, while recording inverse reasoning traces as training supervision. Given a detailed image annotation $P_{\mathrm{fine}}$, we first use an LLM to classify it into one of four image generation categories: General, Portrait, Text, and Complex Text. This task-aware classification ensures that the subsequent degradation process is semantically grounded and adapted to the characteristics of each prompt type. Based on the predicted category, we sample a set of applicable degradation strategies $\mathcal{S}$ from a predefined strategy pool.

To approximate the long-tail distribution of real-world user inputs, we introduce stochasticity into the degradation process. Specifically, a subset of strategies is sampled from $\mathcal{S}$ according to predefined probability distributions. These strategies include stylistic simplification, colloquialization, and removal or underspecification of visual details such as lighting, texture, layout, and background. Applying them to $P_{\mathrm{fine}}$ produces a degraded prompt $P_{\mathrm{short}}$. By adjusting the sampling proportions, the pipeline generates training examples with varying difficulty, ambiguity, and information density.

This construction naturally yields an inverse reasoning chain, \textit{i.e.}, a Chain-of-thought (CoT) for prompt enhancement. Since each degradation operation $s \in \mathcal{S}$ removes or obscures information from the original annotation, its reverse defines a principled trajectory for prompt recovery and enrichment. The resulting triplet $(P_{\mathrm{short}}, \mathrm{CoT}, P_{\mathrm{fine}})$ allows the model to learn both the enhanced prompt and the underlying intent-expansion process, such as inferring lighting, material, spatial, and stylistic cues from the remaining attributes. This reverse-engineering pipeline is used for T2I generation tasks. For image editing, where the input image already provides rich visual context, we instead use an MLLM to summarize long-form annotations into concise editing prompts, avoiding unnecessary stochastic degradation.

\paragraph{PE Training}
The PE module is initialized from Qwen3.5-9B~\citep{qwen35blog} and trained as a unified prompt enhancement model for both image generation and image editing. The training process consists of two consecutive stages: SFT followed by RL. This two-stage design first equips the model with stable rewriting behavior from curated supervision, and then further aligns the rewritten prompts with downstream image generation quality.

During SFT, the model is trained on the constructed dataset with the standard next-token prediction objective, learning prompt enhancement capabilities for intent preservation, scene enrichment, and compositional organization across both generation and editing scenarios. While generation prompts require richer visual elaboration, editing prompts demand faithful instruction preservation and sensitivity to the existing visual context. Since SFT relies on static textual references and cannot directly optimize downstream image quality, we further introduce an RL stage based on GRPO~\citep{shao2024deepseekmath}. The PE model generates candidate enhanced prompts, which are fed into a frozen image generator, and is optimized with rewards combining MLLM-based visual consistency, MLLM-based aesthetic quality, and rule-based textual constraints. This end-to-end training encourages rewrites that better align with user intent while improving the visual outcomes of the generated images.

\begin{figure}[H]
  \centering
  \includegraphics[width=0.9\linewidth]{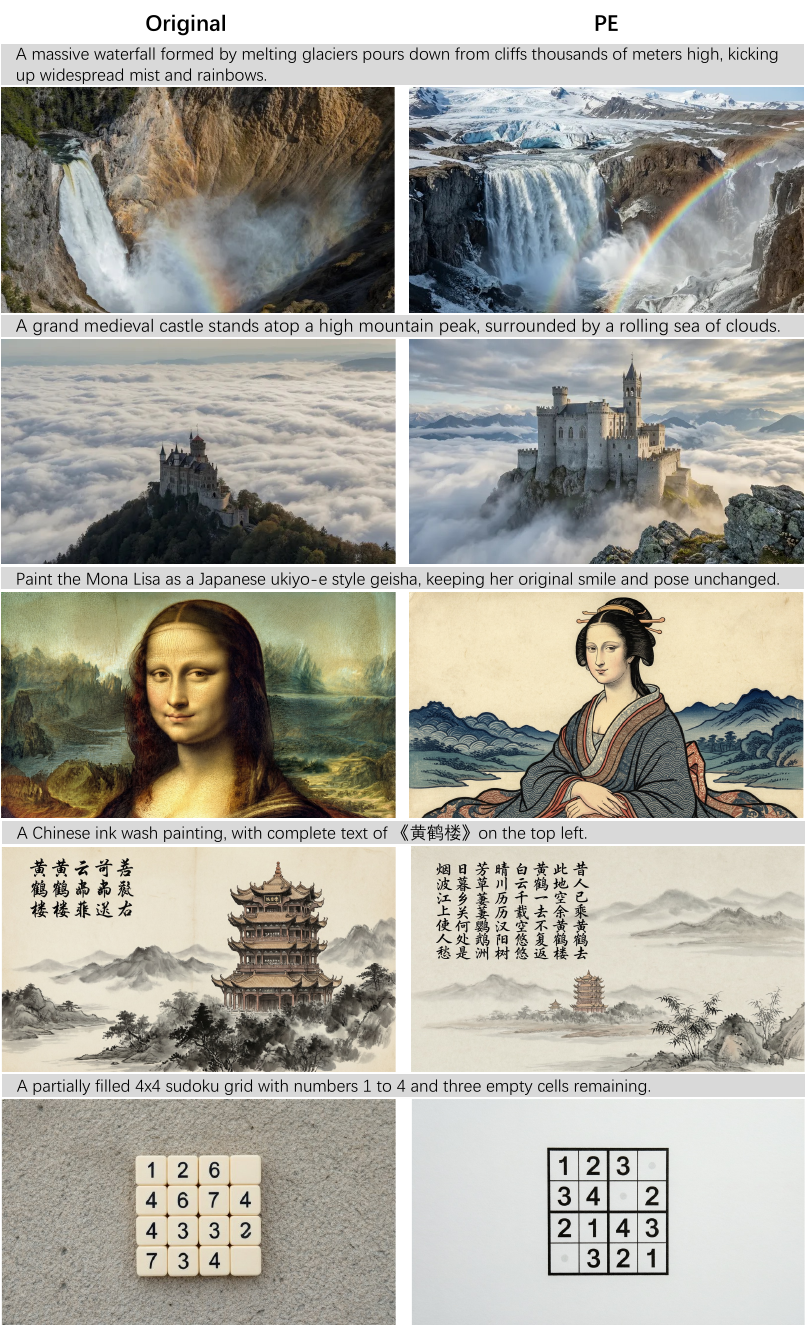} 
  \caption{
      Qualitative comparison of T2I results using the original captions and prompt-enhanced captions.
  }
  \label{fig:pe}
\end{figure}

By combining supervised rewriting objectives with generation-aware reinforcement learning, the PE module is grounded in both textual supervision and downstream visual feedback. As a result, it produces enhanced prompts that are more faithful, expressive, and effective for image generation and editing. As illustrated in Figure~\ref{fig:pe}, the PE module consistently improves generation quality, prompt following, and reasoning performance.

\section{Training}

\subsection{Multistage Training}
During training, we employ a multistage training strategy comprising three phases: pre-training, continual pre-training, and supervised fine-tuning. Across these stages, we progressively adjust the image resolution, data filtering criteria, and data composition, enabling the model to evolve from learning fundamental semantic representations to modeling fine-grained visual details. The detailed configurations are summarized in Table.~\ref{tab:config}.

\begin{table*}[h]
\centering
\caption{Training configurations, data distribution, and hyperparameters used in our experiments.}
\label{tab:config}
\begin{tabular}{lccc}
\toprule
\textbf{Configuration} & \textbf{Pre-training} & \textbf{Continual Pre-training} & \textbf{Supervised Fine-tuning} \\
\midrule

\multicolumn{4}{l}{\textbf{Training Process}} \\
Steps (K)        & 700 & 250 & 10 \\
Resolution       & 256/512 & 512/1024/2048 & 512/1024/2048 \\
Batch Size (K)   & 32/16 & 16/8/4 & 16/8/4 \\
\midrule

\multicolumn{4}{l}{\textbf{Data Distribution}} \\
Type             & T2I/TI2I & T2I/TI2I & T2I/TI2I \\
Ratio            & 0.9/0.1 & 0.7/0.3 & 0.7/0.3 \\
\midrule

\multicolumn{4}{l}{\textbf{Hyperparameters}} \\
Optimizer        & Adam & Adam & Adam \\
Weight Decay     & 0.001 & 0.001 & 0.001 \\
Grad. Norm Clip  & 1.0 & 1.0 & 1.0 \\
Uncond. Dropout  & 0.1 & 0.1 & 0.1 \\
Learning Rate    & $1\times10^{-4}$ & $2\times10^{-5}$ & $1\times10^{-5}$ \\
\bottomrule
\end{tabular}
\end{table*}

\paragraph{Pre-training}
In the pre-training stage, the model primarily learns basic semantic representations. We train the model for 700K steps at relatively low resolutions to improve data throughput. The training data consists of a 9:1 mixture of T2I and TI2I data. The learning rate is set to $1\times10^{-4}$, allowing the model to learn robust and general-purpose visual representations from large-scale image-text data.

\paragraph{Continual pre-training}
In the continual pre-training stage, the model further improves generation quality and adapts to higher-resolution inputs. The model is trained for 250K steps, while the image resolution is gradually increased to 512--2048 to better capture fine-grained visual details. The data distribution is adjusted to a 7:3 mixture of T2I and TI2I data, strengthening image editing capabilities while maintaining strong text-to-image generation performance. The learning rate is reduced to $2\times10^{-5}$ to ensure stable optimization during this stage.

\paragraph{Supervised fine-tuning}
In the supervised fine-tuning stage, we focus on improving the aesthetic quality of generated images. The model is trained for approximately 10K steps. To enhance fine-grained visual details while preserving the model's world knowledge, the learning rate is further reduced to $1\times10^{-5}$. For the training data, we sample from diverse data categories and apply strict filtering together with manual curation to ensure high aesthetic quality.

\begin{figure}[p]
    \makebox[\linewidth]{
        \includegraphics[width=\linewidth]{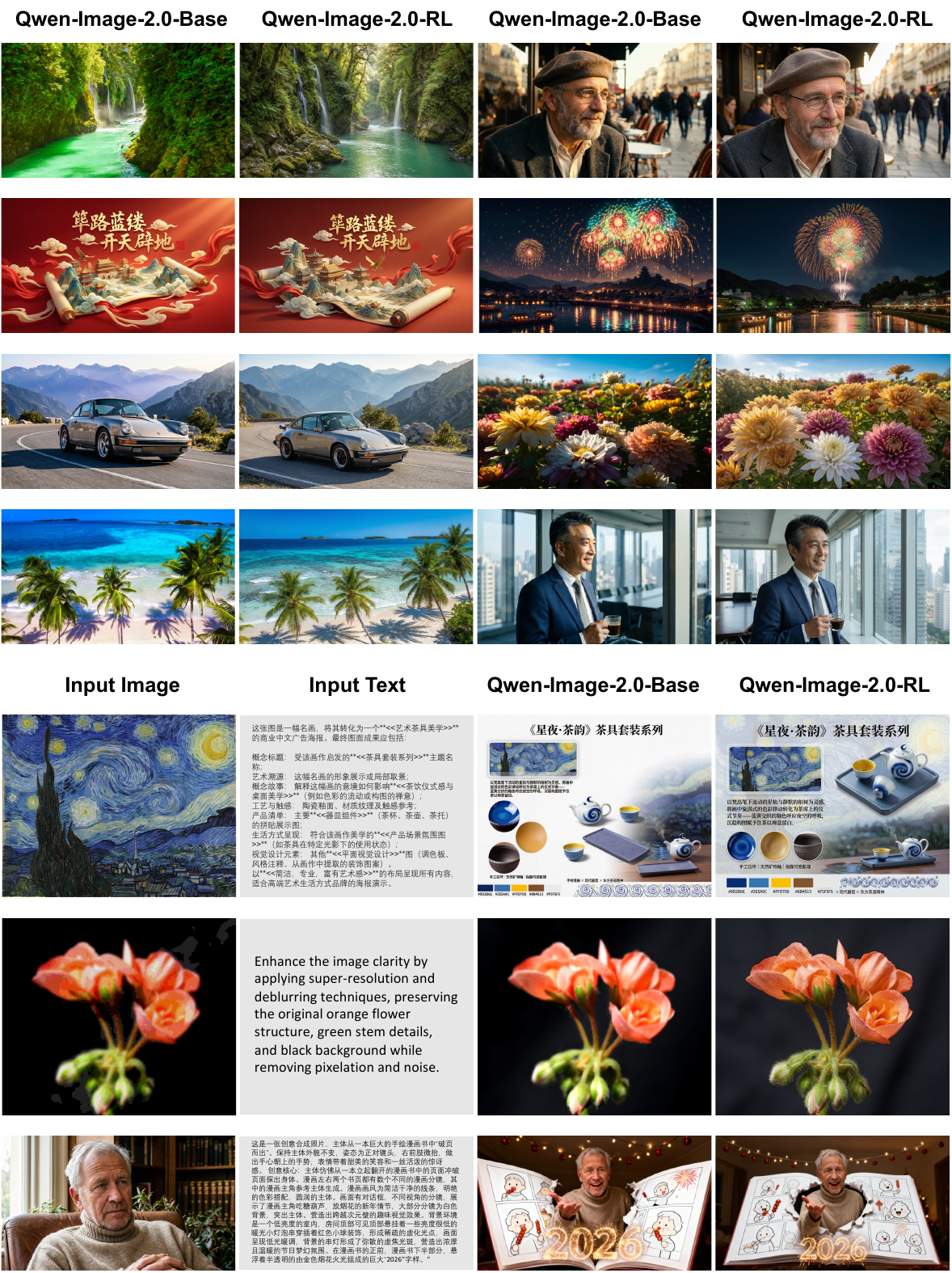}
    }
    \caption{Qualitative comparison between Qwen-Image-2.0-Base and Qwen-Image-2.0-RL across various T2I and TI2I scenarios. Qwen-Image-2.0-RL further improves the visual quality of Qwen-Image-2.0-Base in diverse scenarios, including portraits, landscapes, posters, and natural scenes.}
    \label{fig:RL_comparison}
\end{figure}

\subsection{Reinforcement Learning with Human Feedback}
To align Qwen-Image 2.0 more closely with human preferences and to enhance generation quality across both T2I and TI2I tasks, we develop an RLHF pipeline that refines the base diffusion model through multi-dimensional reward signals and a sample-efficient optimization algorithm. This procedure yields consistent improvements in perceptual quality and task-specific controllability.

\paragraph{Reward modeling} We construct task-specific composite reward models from distinct human preference annotation datasets, with each model targeting a particular evaluation dimension:

\begin{itemize}
        \item \textbf{Aesthetic reward (for T2I).} Assesses the intrinsic visual quality of generated images, emphasizing compositional balance, realistic illumination, texture fidelity, and overall artistic coherence.
        \item \textbf{Image-text alignment reward (for T2I).} Measures semantic correspondence between the generated image and the input prompt, explicitly penalizing outputs that omit, misinterpret, or contradict user-specified requirements.
        \item \textbf{Portrait reward (for T2I).} Provides a specialized optimization signal for human-subject generation, improving anatomical plausibility, facial proportion accuracy, identity-preserving facial details, and fine-grained skin and hair texture realism.
    
        \item \textbf{Instruction-following reward (for TI2I).} Evaluates whether user-specified modifications are accurately executed, covering editing operations such as object replacement and style transfer.
        \item \textbf{Visual consistency reward (for TI2I).} Preserves the identity and structural integrity of unmodified regions by enforcing strict consistency in geometric layout, spatial topology, and semantic features between the source and edited images.
\end{itemize}

All reward models are calibrated to operate on comparable scales, and their weights are dynamically adjusted throughout training to avoid over-optimization toward any single dimension.

\paragraph{Training} We optimize the base diffusion model using an adapted GRPO framework~\citep{liu2025flowgrpo,wang2025grpo,zheng2025diffusionnft}. A key design consideration in diffusion-based reinforcement learning is whether Classifier-free Guidance (CFG, \citealt{ho2022classifier}) should be employed during rollout sampling and policy optimization. Existing studies adopt divergent strategies: some methods apply CFG in both rollout and training stages~\citep{liu2025flowgrpo,wang2025grpo}, whereas others omit it entirely~\citep{zheng2025diffusionnft}. In our RLHF pipeline, we adopt a hybrid strategy: CFG is used during rollout sampling to generate high-quality candidates for reward evaluation, while the unconditional branch is excluded from the policy optimization objective. This design preserves the visual fidelity and structural coherence of sampled images, thereby providing more reliable reward signals, while substantially reducing the computational overhead associated with optimizing the unconditional model. The resulting RL-aligned model is denoted as \textbf{Qwen-Image-2.0-RL}. In practice, we further refine the optimization process by dynamically adjusting the prompt distribution across tasks and calibrating the relative weights of individual reward models, leading to improved final visual quality.

\paragraph{Results} Qualitative evaluations indicate that the proposed RLHF pipeline produces consistent gains across both T2I generation and image editing tasks. For T2I generation, Qwen-Image-2.0-RL demonstrates notable improvements in texture fidelity and overall image realism. In image editing scenarios, Qwen-Image-2.0-RL likewise enhances texture quality and visual consistency. Figure~\ref{fig:RL_comparison} presents side-by-side comparisons of T2I and editing outputs before and after RL alignment, illustrating the resulting improvements in visual refinement.

\subsection{Few-step Distillation}
We aim to distill our multi-step model into a few-step variant that is more efficient, while preserving visual quality and prompt-following ability. However, due to the architectural complexity of large multimodal models, such distillation remains highly challenging, especially when the goal is to retain the model’s full capabilities across diverse scenarios, such as portrait generation, landscape synthesis, and text rendering, under an extremely limited number of function evaluations (NFEs). 

Recent advances in diffusion distillation have explored a broad spectrum of techniques, including trajectory-based optimization~\citep{song2023consistency, lu2024simplifying, geng2025mean} and distribution-level matching~\citep{sauer2024adversarial, sauer2024fast, liu2025decoupled, wu2026diversity}. However, most existing studies are confined to class-conditional settings, predominantly on ImageNet~\citep{deng2009imagenet}, leaving their efficacy in broader and more practically relevant scenarios, including T2I generation and image editing, largely underexplored. Among advanced diffusion distillation paradigms, we employ Distribution Matching Distillation (DMD; \citealt{yin2024one, yin2024improved}), motivated by its strong empirical stability and consistent effectiveness on heterogeneous visual generative architectures (\emph{e.g.}, Stable Diffusion, \citealt{rombach2021highresolution}), as well as its demonstrated versatility in diverse generation scenarios.

Concretely, given a conditional few-step student generator
$G_{\bm{\theta}}$ parameterized by $\bm{\theta}$, an initial Gaussian noise vector $\bm{\epsilon}\sim\mathcal{N}(\mathbf{0},\mathbf{I})$, and a condition $\bm c\sim p(\bm c)$, we denote the corresponding clean-state prediction as $\bm{x}_{\bm{\theta}} = G_{\bm{\theta}}(\bm{\epsilon}, \bm c).$ Here, $G_{\bm{\theta}}$ is used broadly: $\bm{x}_{\bm{\theta}}$ may be the final clean sample obtained after the full few-step student trajectory, or a clean state directly predicted from an intermediate student state conditioned on $\bm c$. The gradient of the DMD objective $\ell_{\text{DMD}}(\bm{\theta})$ with respect to the student parameters $\bm{\theta}$ is then given by
\begin{equation}
\nabla_{\bm \theta} \ell_{\text{DMD}}(\bm\theta)
=
\mathbb{E}_{\bm c\sim p(\bm c),\,
\bm{\epsilon}\sim\mathcal{N}(\mathbf{0}, \mathbf{I}),\,
\bm{\xi}\sim\mathcal{N}(\mathbf{0}, \mathbf{I}),\,
t\sim p(t)}
\Big[
\big(
\bm s_{\text{fake}}(\bm{x}_t, t, \bm c)
-
\bm s_{\text{real}}(\bm{x}_t, t, \bm c)
\big)
\nabla_{\bm\theta} \bm{x}_{\bm{\theta}}
\Big],
\end{equation}
where $\bm{\xi}$ denotes an independent Gaussian noise vector, 
$t\in[0,1]$ is the diffusion time sampled from a prescribed distribution 
$p(t)$ (\emph{e.g.}, a logit-normal distribution), and $\bm{x}_t$ is obtained by linearly interpolating between the conditionally generated clean sample 
$\bm{x}_{\bm{\theta}}$ and the noise vector $\bm{\xi}$:
\begin{equation}
\bm{x}_t=(1-t)\bm{x}_{\bm{\theta}}+t\bm{\xi}.
\end{equation}
Here,
$\bm s_{\text{fake}}(\bm{x}_t, t, \bm c)
=
\nabla_{\bm{x}_t}\log p_{\text{fake}, t}(\bm{x}_t \mid \bm c)$
denotes the conditional score function associated with the student-induced distribution at noise level $t$; in practice, this score is estimated by an auxiliary fake score model trained on conditionally generated student samples using a flow-matching objective.
Meanwhile,
$\bm s_{\text{real}}(\bm{x}_t, t, \bm c)
=
\nabla_{\bm{x}_t}\log p_{\text{real}, t}(\bm{x}_t \mid \bm c)$
denotes the conditional target score provided by the pretrained teacher diffusion model at the same noise level.

\begin{figure}[t]
\centering
\includegraphics[width=1\linewidth]{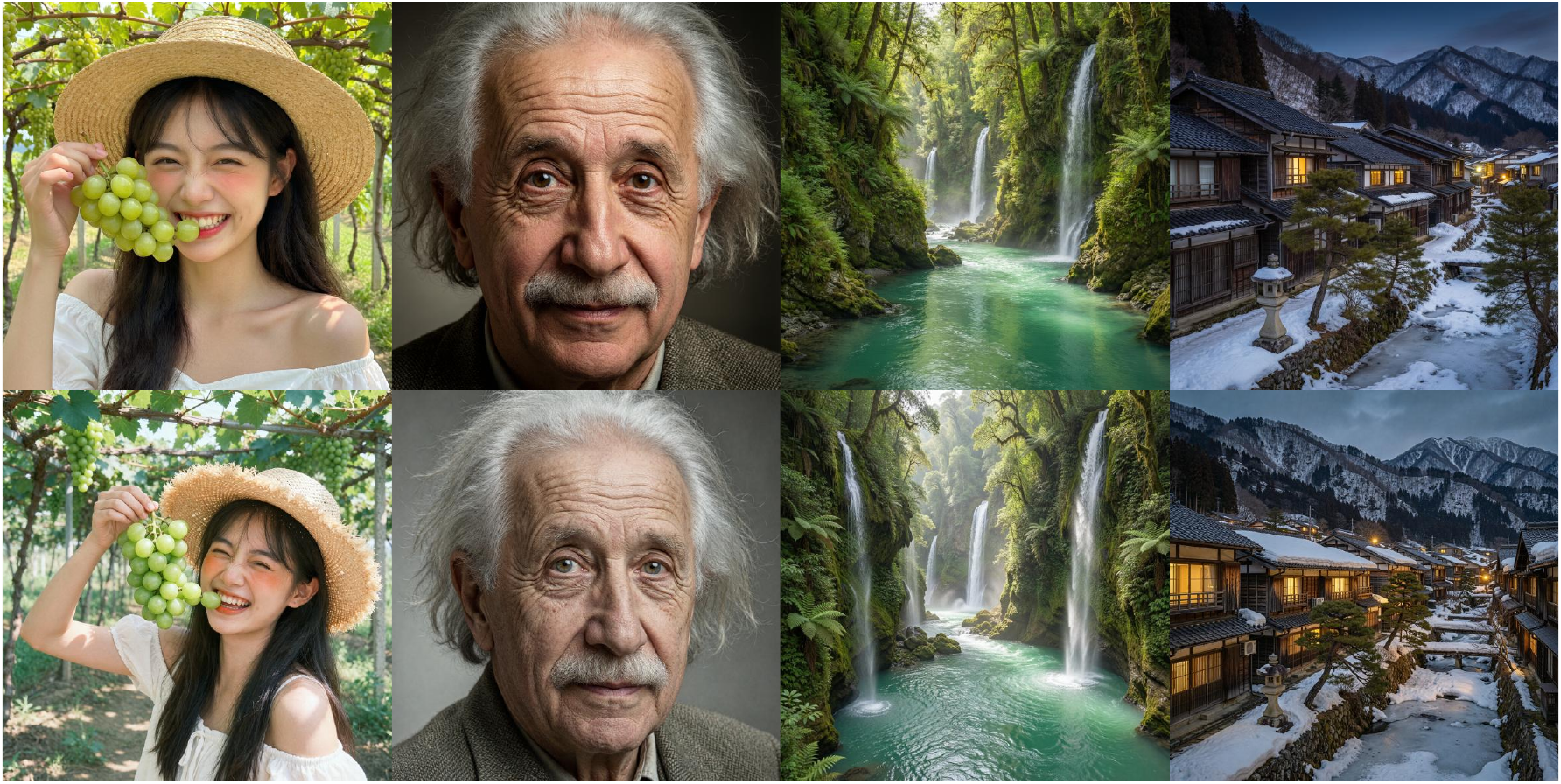}
   \caption{Qualitative comparison between the multi-step teacher and the few-step distilled student. The top row shows images generated by Qwen-Image-2.0-RL with 40 sampling steps, while the bottom row shows images generated by Qwen-Image-2.0-Distillation with only 4 NFEs. Across diverse prompts, including portraits, landscapes, and natural scenes, the 4-NFE student preserves visual quality, semantic alignment, and compositional coherence comparable to the 40-step teacher, while reducing inference cost.}
\label{fig:distill_case}
\end{figure}

\paragraph{Results}
Starting from Qwen-Image-2.0-Base as the multi-step teacher, we apply the above distillation procedure to obtain \textbf{Qwen-Image-2.0-Distillation} as the few-step student optimized for efficient inference. As shown in Figure~\ref{fig:distill_case}, the distilled $4$-NFE student produces results visually comparable to the $40$-step teacher across diverse prompts and visual domains. It preserves detailed appearance, coherent composition, and faithful semantic alignment, while substantially reducing the number of function evaluations. These comparisons show that our DMD-based distillation effectively compresses the sampling trajectory while maintaining perceptual quality and prompt-following capability.

\section{Benchmark and Qualitative Evaluation}
\label{sec:exp}

\subsection{LMArena Benchmark Evaluation}
To assess the image generation capability of Qwen-Image-2.0, we evaluate it on LMArena~\citep{arena_ai_leaderboard}, a leading benchmark grounded in real-world user preferences. On the T2I leaderboard, users anonymously compare images produced by different models from the same prompt, without knowing the identity of the generation model. This blind evaluation protocol promotes fairness, while the ELO-based ranking system offers a preference-oriented measure of model performance.

\begin{figure}[t]
    \makebox[\linewidth]{
        \includegraphics[width=\linewidth]{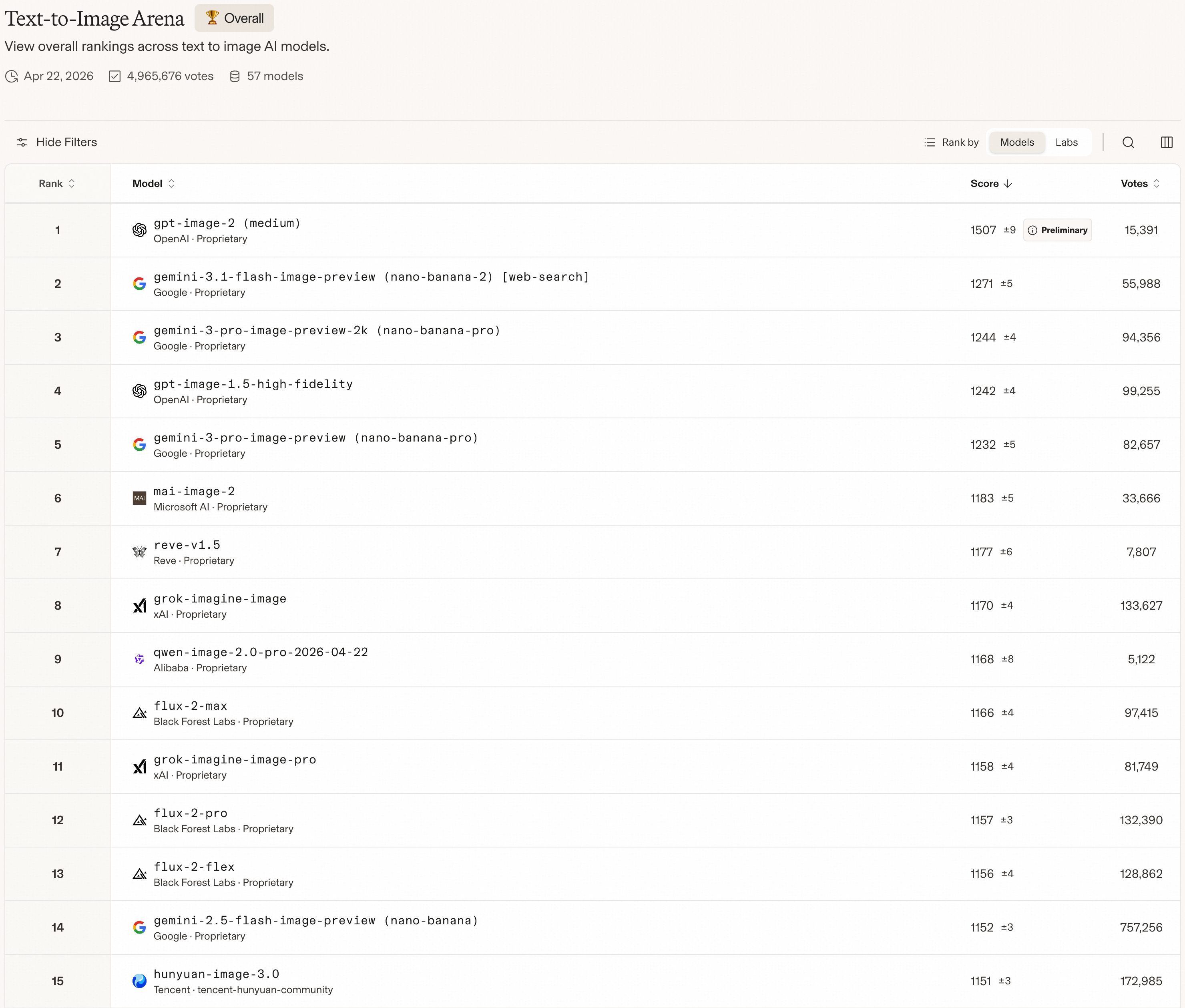}
    }
    \caption{Results from LMArena (accessed April 22, 2026).}
    \label{fig:lmarena_screenshot}
\end{figure}

As shown in Figure~\ref{fig:lmarena_screenshot}, Qwen-Image-2.0 achieves strong performance on this widely recognized image generation benchmark, ranking \#9 globally and \#1 among Chinese models. In direct comparison with leading international models, Qwen-Image-2.0 reaches the top tier with an ELO score of 1168 and outperforms Nano Banana. As shown in Figure~\ref{fig:lmarena_radar}, Qwen-Image-2.0 delivers substantial improvements over previous Qwen-Image series models in both image generation and editing, demonstrating clear advances in overall visual quality, editing capability, and practical usability.

\subsection{Qualitative Results on Text-to-image Generation}
\begin{figure}[t]
  \centering
  \includegraphics[width=\linewidth]{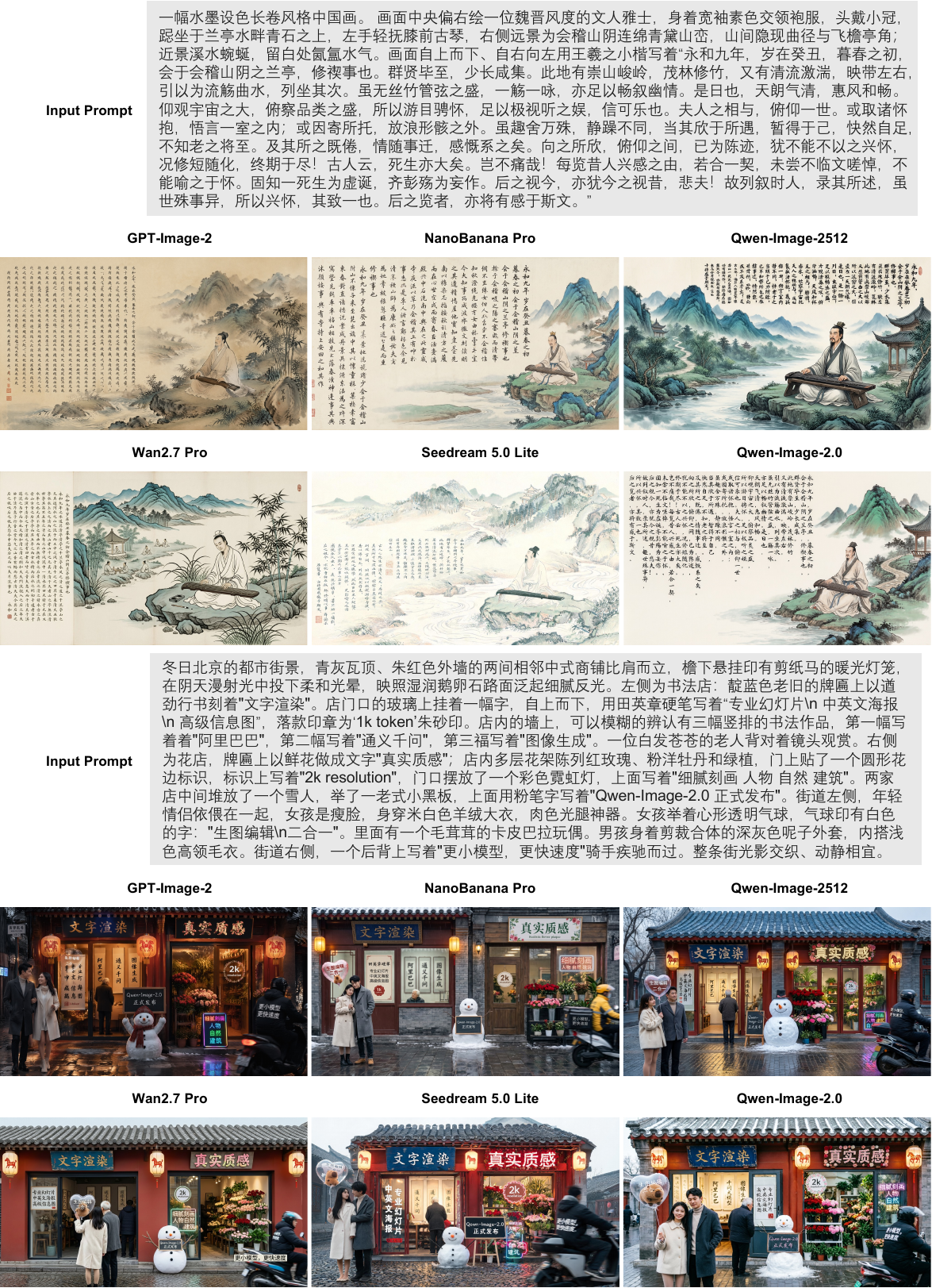} 
  \caption{
      Qualitative comparison of text rendering results.
  }
  \label{fig:text_rendering}
\end{figure}

\begin{figure}[t]
  \centering
  \includegraphics[width=\linewidth]{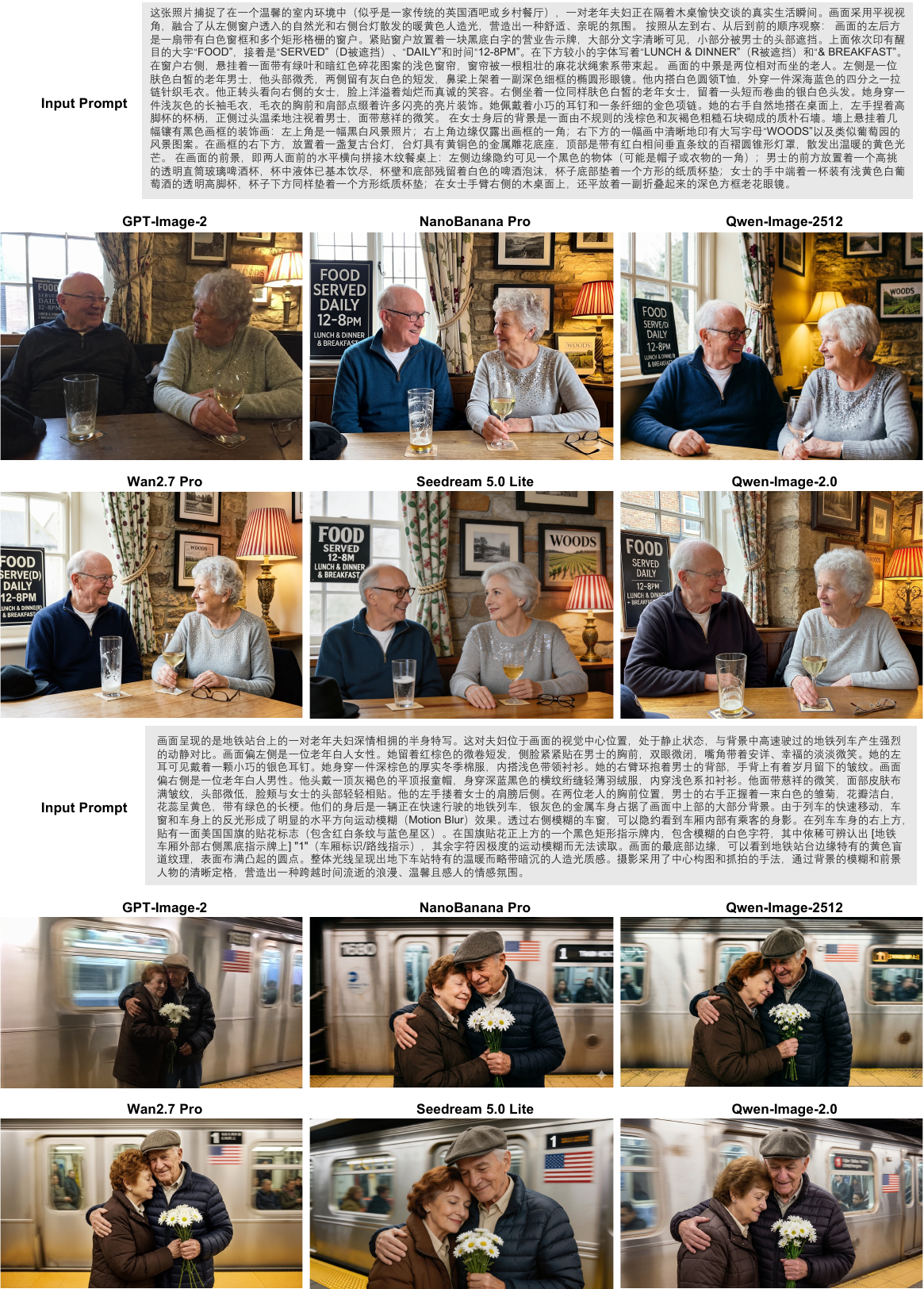} 
  \caption{
      Qualitative comparison of portrait generation results.
  }
  \label{fig:portrait}
\end{figure}

\begin{figure}[t]
  \centering
  \includegraphics[width=\linewidth]{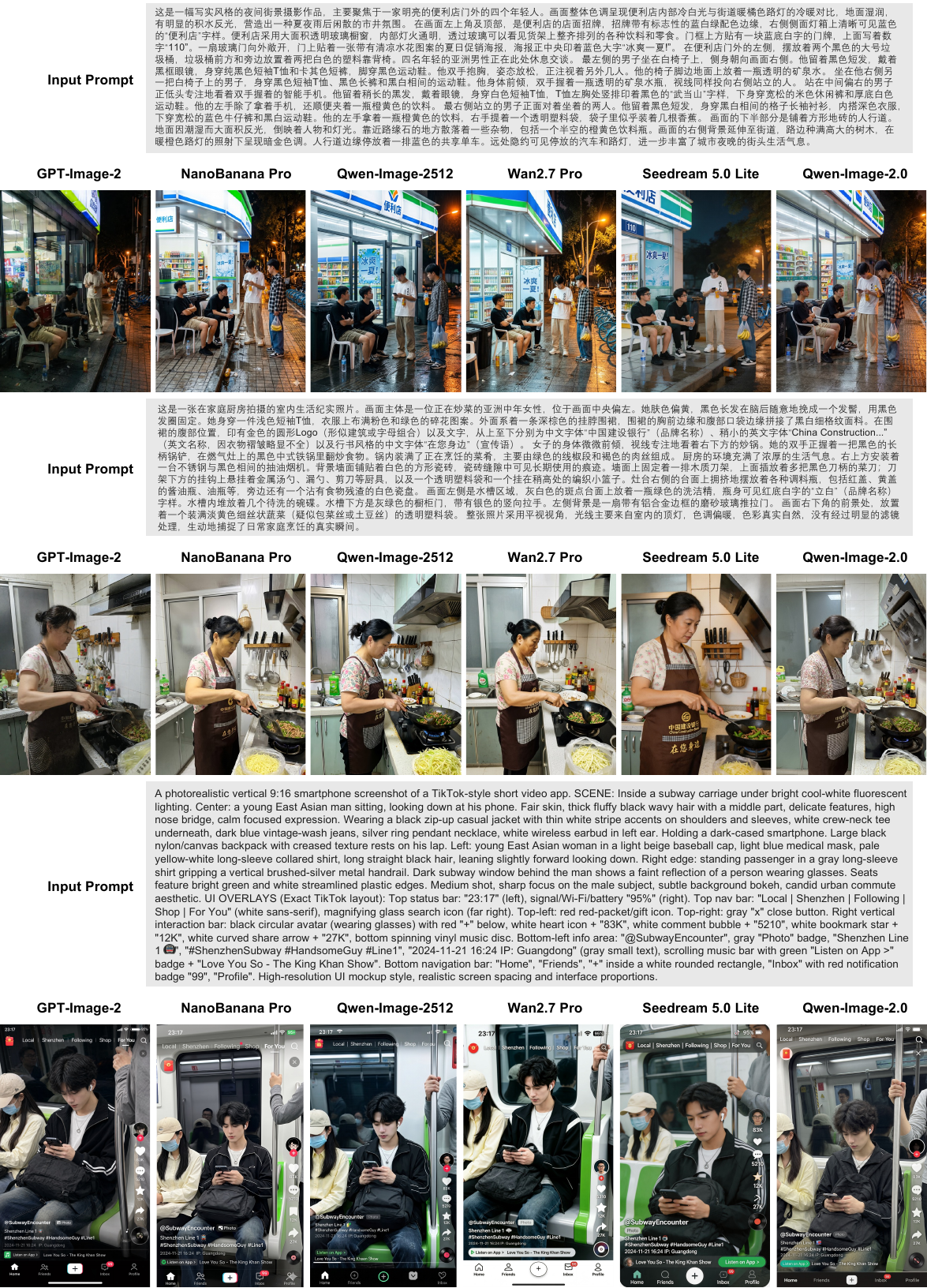} 
  \caption{
      Qualitative comparison of portrait generation results.
  }
  \label{fig:portrait2}
\end{figure}

\begin{figure}[t]
  \centering
  \includegraphics[width=\linewidth]{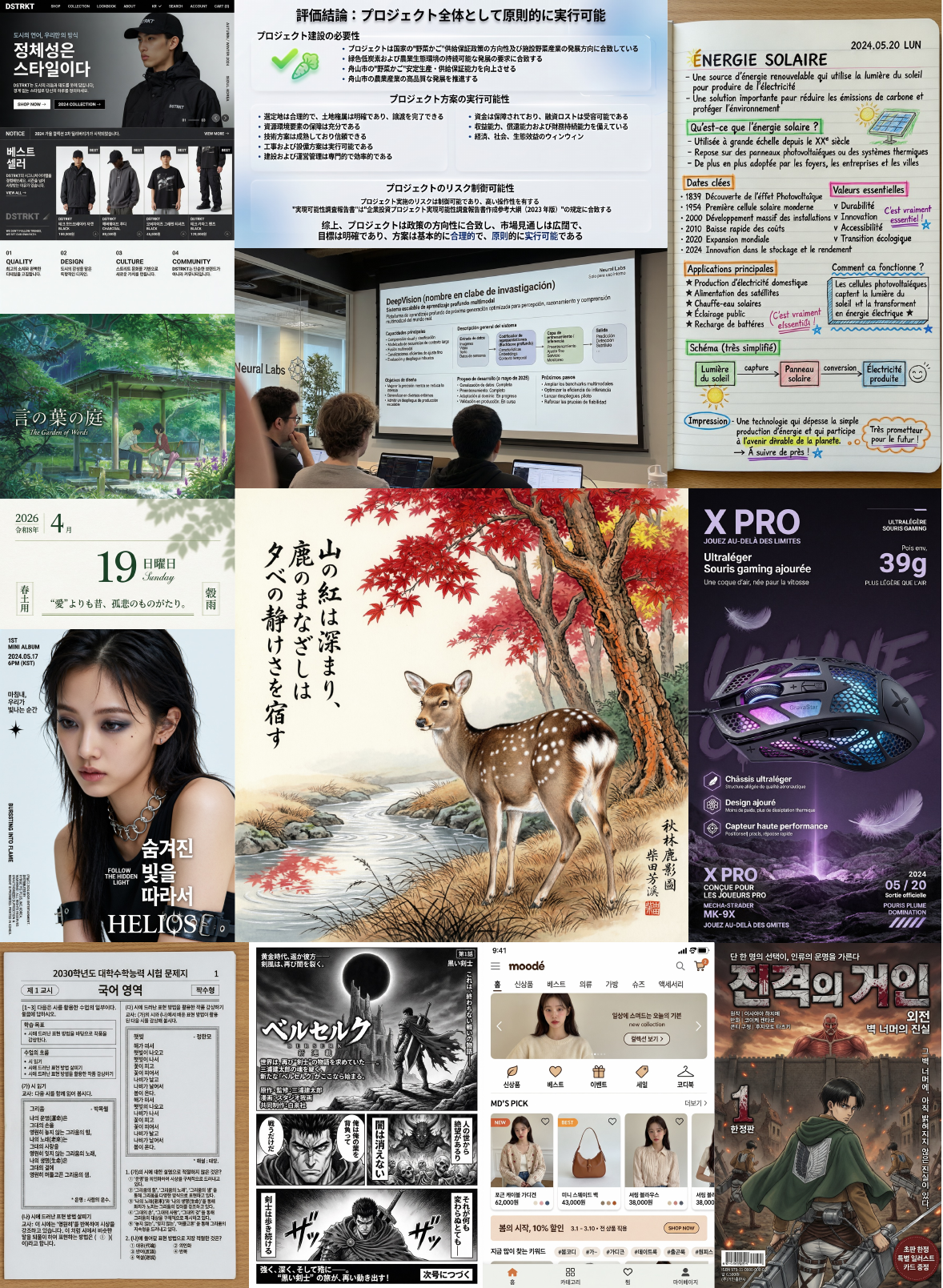} 
  \caption{
      Visualization of multilingual rendering by Qwen-Image-2.0.
  }
  \label{fig:multilingual}
\end{figure}

\begin{figure}[t]
  \centering
  \includegraphics[width=\linewidth]{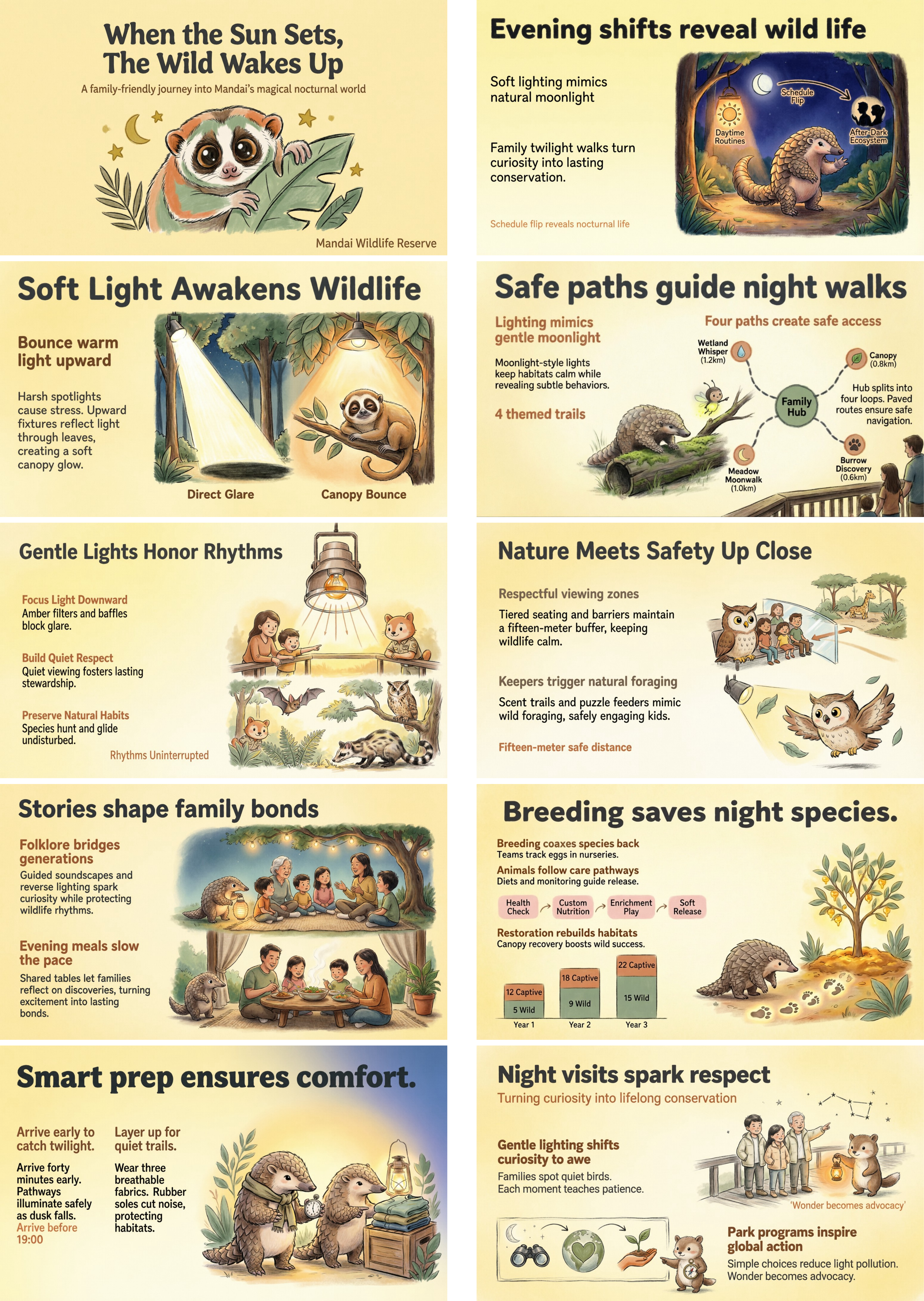} 
  \caption{
      Visualization of slide generation by Qwen-Image-2.0.
  }
  \label{fig:slide}
\end{figure}

We qualitatively evaluate Qwen-Image-2.0 on T2I generation, covering text rendering (Figure~\ref{fig:text_rendering}), portrait generation (Figures~\ref{fig:portrait} and~\ref{fig:portrait2}), multilingual text rendering (Figure~\ref{fig:multilingual}), and slide generation (Figure~\ref{fig:slide}).

\paragraph{Text rendering}
Figure~\ref{fig:text_rendering} presents a qualitative comparison of Chinese text rendering across different models. In the first example, GPT-Image-2 renders the characters at an excessively small scale and introduces frequent character-level errors; NanoBanana Pro fails to reproduce the complete prompt sequence, erroneously duplicating certain segments while also introducing multiple typos; Qwen-Image-2512 exhibits inconsistent font sizing and numerous miswritten characters; Wan2.7 Pro disregards the specified textual prompt entirely, generating a substantial amount of unrelated content instead; and Seedream 5.0 Lite produces undersized, poorly legible text that is further compromised by frequent character inaccuracies. In contrast, only Qwen-Image-2.0 successfully fulfills the text-rendering objective with negligible errors, while ensuring that the generated typographic style is harmoniously integrated with the overall visual composition. In the second example, GPT-Image-2 produces largely illegible gibberish on the vertical posters and small details despite rendering the main headers; NanoBanana Pro hallucinates incoherent text on the left poster; Qwen-Image-2512 generates unreadable character on the side posters; Wan2.7 Pro correctly renders the shop signboards but fails to spatially bind the rider's back text, instead outputting the phrase as a detached subtitle-style overlay at the bottom right rather than integrating it onto the rider's garment, thereby disrupting the scene's physical realism; and Seedream 5.0 Lite renders the main signs but introduces erroneous and disjointed characters on the vertical banners. Remarkably, Qwen-Image-2.0 uniquely preserves character-level accuracy, correct spatial binding for all text elements, and a coherent, physically grounded scene composition.

\paragraph{Portrait generation}
Figure~\ref{fig:portrait} presents a qualitative comparison of portrait generation across different models. In the first example, GPT-Image-2 renders the background stone wall with an overly smooth and artificial texture, lacking the rustic irregularity and material realism expected of a traditional interior; Qwen-Image-2512 and Wan2.7 Pro misinterpret the occlusion instruction by literally rendering the text as ``SERVE(D)''; Seedream 5.0 Lite omits the word ``DAILY'' entirely and produces the garbled time ``12-8M''; and NanoBanana Pro, although capturing the main headers, renders the signboard as a flat and unnatural overlay that lacks physical integration with the window frame. In contrast, Qwen-Image-2.0 is the only model that simultaneously achieves high-fidelity text rendering on the signboard while preserving a photorealistic atmosphere through accurate material textures and natural lighting consistency. In the second example, NanoBanana Pro hallucinates large and incorrect numbers (``1680'') directly on the train body, violating the textual constraints specified in the prompt; Qwen-Image-2512 fails to apply the required extreme motion blur to the signboard, leaving text unnaturally distorted; Wan2.7 Pro mistakenly renders Chinese on the train; and Seedream 5.0 Lite not only produces overly smooth hair and skin textures, but also renders the numeral ``1'' perfectly legible and thereby disrupting the physical realism. By comparison, Qwen-Image-2.0 can successfully generate strong horizontal motion blur on the train and correctly position the American flag decal, while preserving the warm artificial lighting and intimate emotional focus on the couple.

\subsection{Qualitative Results on Image Editing}
\begin{figure}[t]
  \centering
  \includegraphics[width=0.9\linewidth]{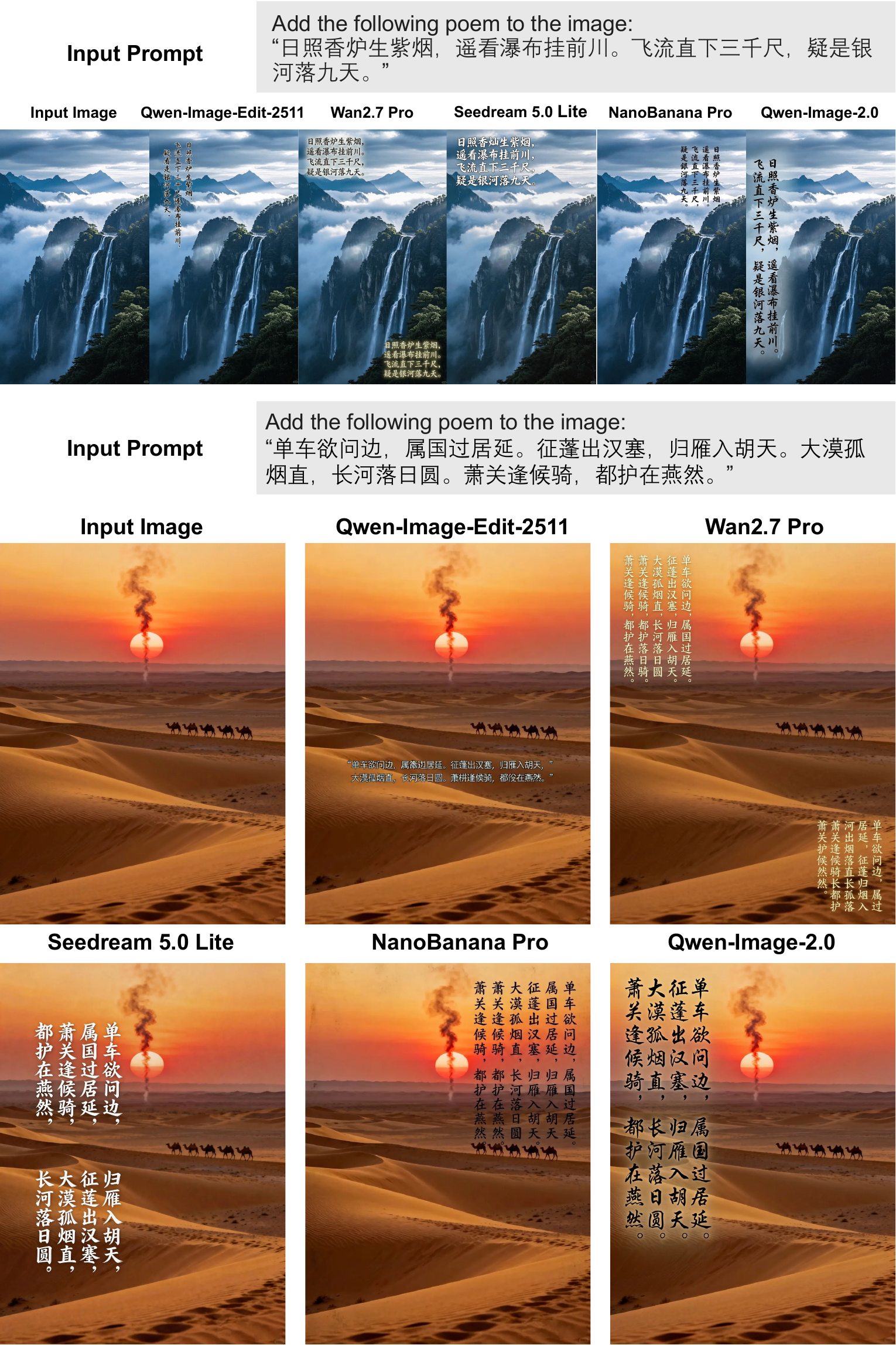} 
  \caption{
      Qualitative comparison of complex Chinese text rendering in an image editing task. Qwen-Image-2.0 demonstrates superior accuracy and aesthetic quality, and is the only model capable of rendering classical Chinese poetry both accurately and aesthetically.
  }
  \label{fig:text_edit}
\end{figure}

\begin{figure}[t]
  \centering
  \includegraphics[width=0.95\linewidth]{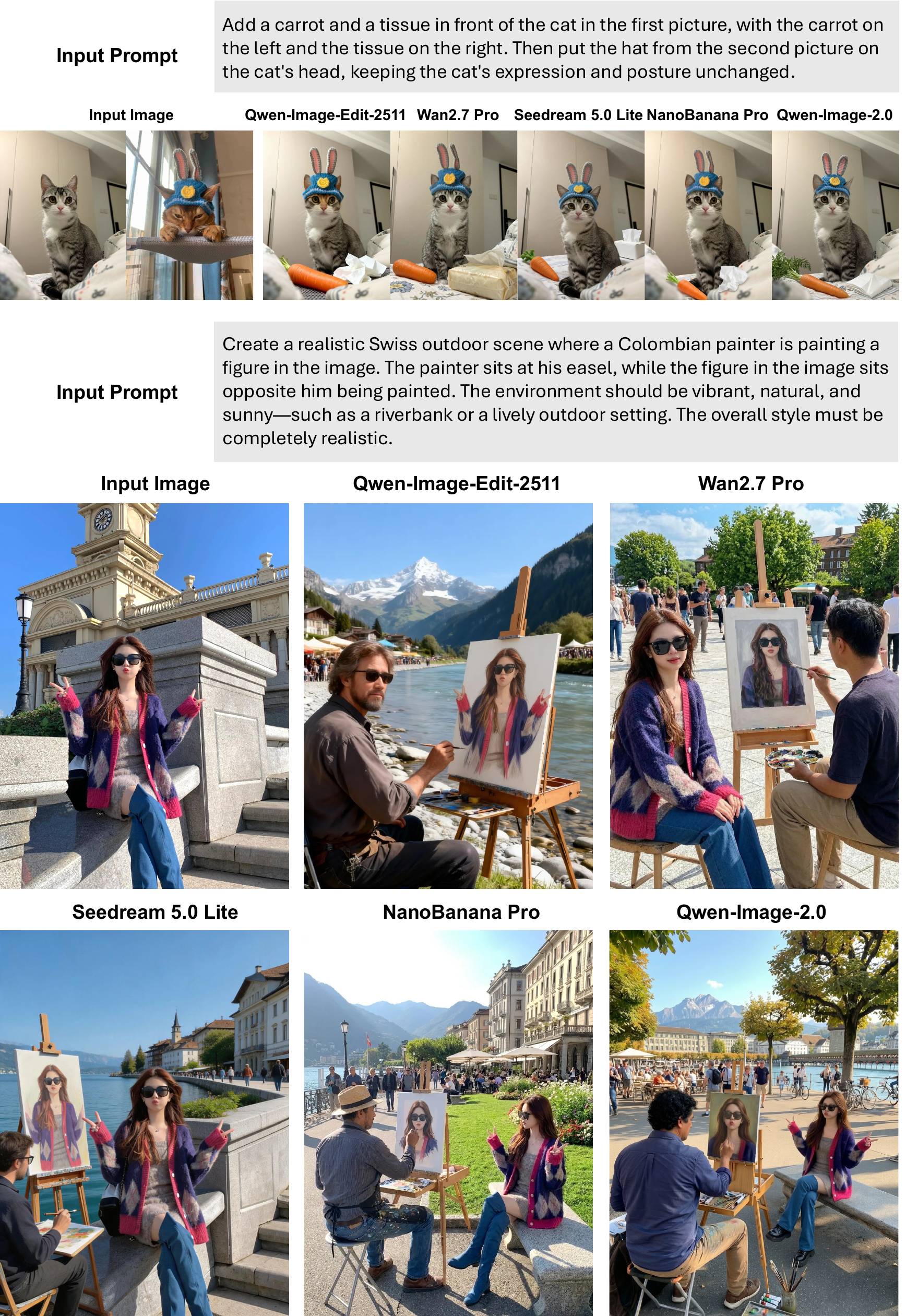} 
  \caption{
      Qualitative comparison of identity preservation. In both single-image and multi-image editing tasks, Qwen-Image closely follows user instructions while maintaining fine-grained object details, including facial expressions, posture, and overall appearance. These results highlight its strong capability for precise object-level editing without sacrificing visual consistency.
  }
  \label{fig:id_consistency}
\end{figure}

For TI2I editing, we evaluate Qwen-Image-2.0 on complex Chinese text rendering and identity preservation across single-image and multi-image editing tasks, with examples shown in Figures~\ref{fig:text_edit} and~\ref{fig:id_consistency}.

\paragraph{Complex text rendering}
Figure~\ref{fig:text_edit} presents a qualitative comparison of complex Chinese text rendering across different models. In the first example, Qwen-Image-Edit-2511 and NanoBanana Pro render the characters at an excessively small scale, thereby disrupting the visual balance with the landscape; Wan2.7 Pro erroneously duplicates the poem by rendering two separate copies within the same image; and Seedream 5.0 Lite exhibits a character-level error, miswriting one character in the opening line of the poem. In contrast, only Qwen-Image-2.0 produces a layout consistent with the traditional ti-hua-shi (poem-on-painting) aesthetic, featuring an appropriate font scale, vertical right-to-left orientation, and harmonious placement within the negative space of the sky, while simultaneously preserving character-level accuracy. In the second example, which contains a longer 40-character poem with multiple rare and structurally complex characters, the baseline models exhibit clear failures: NanoBanana Pro reorders the couplets, disrupting the canonical line sequence of the poem; Seedream 5.0 Lite fragments the poem into disjoint columns that break the original reading order; and Qwen-Image-Edit-2511 produces text that is barely legible at the rendered scale. Remarkably, Qwen-Image-2.0 is the only model that simultaneously preserves character-level accuracy, the canonical line order, and a coherent vertical composition.

\paragraph{Identity preservation}
Figure~\ref{fig:id_consistency} provides a qualitative comparison of identity preservation across models on both single-image and multi-image editing tasks. In the first example, the edit requires placing a carrot and a tissue in front of the cat from the first image while transferring the hat from the second image onto its head. The baseline models exhibit evident failures: Qwen-Image-Edit-2511 changes the cat's fur color and pattern; Wan2.7 Pro modifies the cat's original posture; Seedream 5.0 Lite incorrectly places the carrot and tissue behind the cat; and NanoBanana Pro renders the inserted objects with insufficient realism. In contrast, only Qwen-Image-2.0 preserves the cat's identity while accurately satisfying the editing instructions. In the second example, the task is to generate a realistic Swiss outdoor scene in which a Colombian painter paints the figure from the input image. The baseline models again fail in different ways: Qwen-Image-Edit-2511 omits the subject being painted; Wan2.7 Pro changes the painter's ethnicity and produces a female figure that no longer resembles the input; Seedream 5.0 Lite places the easel inconsistently; and NanoBanana Pro renders the subject with substantially different facial features and posture. By comparison, Qwen-Image-2.0 uniquely preserves the subject's facial identity, sunglasses, and distinctive cardigan pattern while correctly composing the multi-element scene, demonstrating strong capability for precise object-level editing without compromising visual consistency.

\clearpage

\section{Conclusion}
In this work, we present \textbf{Qwen-Image-2.0}, a versatile image generation foundation model that supports both T2I generation and instruction-based image editing within a single framework. By combining a strong multimodal encoder, an efficient MMDiT backbone, and a high-compression VAE, Qwen-Image-2.0 addresses several key challenges in real-world image generation, including long-text rendering, multilingual typography, high-resolution photorealism, complex instruction following, and inference efficiency. We hope Qwen-Image-2.0 provides a strong foundation for future research and practical deployment of general-purpose image generation systems.

\section{Authors}

\textbf{Core Contributors\footnote{Alphabetical order.}:} Bing Zhao, Chenfei Wu, Deqing Li, Hao Meng, Jiahao Li, Jie Zhang, Jingren Zhou, Junyang Lin, Kaiyuan Gao, Kuan Cao, Kun Yan, Liang Peng, Lihan Jiang, Niantong Li, Ningyuan Tang, Shengming Yin, Tianhe Wu, Xiao Xu, Xiaoyue Chen, Xihua Wang, Yan Shu, Yanran Zhang, Yi Wang, Yilei Chen, Ying Ba, Yixian Xu, Yujia Wu, Yuxiang Chen, Zecheng Tang, Zekai Zhang, Zhendong Wang, Zihao Liu, Zikai Zhou

\textbf{Contributors\footnote{Alphabetical order.}:} An Yang, Chen Cheng, Chenxu Lv, Dayiheng Liu, Fan Zhou, Hantian Xiong, Hongzhu Shi, Hu Wei, Huihong Zhao, Ivy Liu, Jianwei Zhang, Jiawei Zhang, Kai Chen, Kang He, Levon Xue, Lin Qu, Linhan Tang, Luwen Feng, Minggang Wu, Minmin Sun, Na Ni, Rui Men, Shuai Bai, Sishou Zheng, Tao Lan, Tianqi Zhang, Tingkun Wen, Wei Wang, Weixu Qiao, Weiyi Lu, Wenmeng Zhou, Xiaodong Deng, Xiaoxiao Xu, Xinlei Fang, Xionghui Chen, Yanan Wang, Yang Fan, Yichang Zhang, Yixuan Xu, Yu Wu, Zhiyuan Ma, Zhizhi Cai

\bibliography{colm2024_conference}
\bibliographystyle{colm2024_conference}

\end{document}